\documentclass{article} 
\usepackage{iclr2027_conference,times}

\usepackage[scaled=0.85]{beramono}

\usepackage[utf8]{inputenc} 
\usepackage[T1]{fontenc}    
\usepackage{booktabs}       
\usepackage{amsfonts}       
\usepackage{nicefrac}       
\usepackage{microtype}      

\usepackage{amsmath,amsfonts,bm}

\def\eqref#1{equation~\ref{#1}}

\def\1{\bm{1}}

\DeclareMathAlphabet{\mathsfit}{\encodingdefault}{\sfdefault}{m}{sl}
\SetMathAlphabet{\mathsfit}{bold}{\encodingdefault}{\sfdefault}{bx}{n}

\usepackage{graphicx,xcolor} 

\definecolor{mylinks}{RGB}{41, 171, 226} 
\definecolor{bestred}{RGB}{178, 24, 43}  
\usepackage[colorlinks=true,urlcolor=mylinks,linkcolor=mylinks,citecolor=mylinks
]{hyperref}
\usepackage{natbib}
\usepackage{hyperref}
\usepackage{url}

\usepackage{amsthm}
\usepackage{enumitem}
\usepackage{booktabs}
\usepackage{multirow}
\usepackage{wrapfig}

\usepackage{algorithm}
\usepackage{algpseudocode}
\usepackage[skins,breakable]{tcolorbox}

\newcommand{\sd}[1]{\,\raisebox{-0.25ex}{\tiny$\pm$#1}}
\newcommand{\scell}[2]{#1\sd{#2}}
\newcommand{\sbest}[2]{\textcolor{bestred}{\textbf{#1}\sd{#2}}}
\newcommand{\ssec}[2]{\underline{#1}\sd{#2}}

\newtheoremstyle{gtrlplain}
  {0pt}{0pt}{\itshape}{}{\bfseries}{.}{.5em}{}
\newtheoremstyle{gtrldefn}
  {0pt}{0pt}{}{}{\bfseries}{.}{.5em}{}

\theoremstyle{gtrlplain}

\newtheorem{proposition}{Proposition}

\theoremstyle{gtrldefn}

\newtheorem{example}{Example}
\tcbset{
  thmstyle/.style={
    enhanced, breakable, boxrule=0.7pt, arc=2pt,
    left=8pt, right=8pt, top=6pt, bottom=6pt,
    before skip=10pt, after skip=10pt
  }
}
\tcolorboxenvironment{theorem}    {thmstyle, colframe=blue!55!black,   colback=blue!6}
\tcolorboxenvironment{proposition}{thmstyle, colframe=teal!70!black,   colback=teal!8}
\tcolorboxenvironment{lemma}      {thmstyle, colframe=violet!65!black, colback=violet!7}
\tcolorboxenvironment{corollary}  {thmstyle, colframe=orange!80!black, colback=orange!9}
\tcolorboxenvironment{definition} {thmstyle, colframe=green!45!black,  colback=green!8}
\tcolorboxenvironment{example}    {thmstyle, colframe=purple!60!black, colback=purple!6}
\tcolorboxenvironment{assumption} {thmstyle, colframe=red!55!black,    colback=red!5}

\title{GTRL: Grounding Divide-and-Conquer Value Learning with Temporal Differences}

\author{Abdul Monaf Chowdhury$^{1}$ \quad MD Sameer Iqbal Chowdhury$^{2}$ \\  
\textbf{Shifat E Arman$^{1,3}$ \quad Md Mehedi Hasan$^{1}$} \\
$^{1}$Department of Robotics and Mechatronics Engineering, University of Dhaka, Bangladesh \\
$^{2}$Department of Computer Science, Texas State University, USA \\
$^{3}$Department of Computer Science, University of Oxford, UK
}

\iclrfinalcopy 
\begin{document}

\maketitle
\lhead{Preprint}

\begin{abstract}
In offline goal-conditioned reinforcement learning (GCRL), divide-and-conquer scales to long horizons by joining two shorter segments at a subgoal. However, under stochastic dynamics, the base case of this rule values the luckiest trajectories through the data. The subgoal must also lie on a shared trajectory, so a state-goal pair that no trajectory connects gets no value update at all. To address both, we present \textbf{Grounded Transitive RL (GTRL)}, an offline GCRL value learning algorithm that grounds the divide-and-conquer update with a one-step TD target. Over a single step, TD is correct, as its target averages over the successors and needs no subgoal. GTRL adds this target to the composition rather than replacing it, so every pair receives an update, and the composition still carries the long horizon. GTRL also corrects the bias from hindsight relabeling by reweighting each goal against how reachable it was from other successors. We evaluate our algorithm on nineteen OGBench tasks spanning stochastic, deterministic, and stitching environments, where it achieves the highest average success rate. \\
Code: \url{https://github.com/monaf-chowdhury/gtrl}  
\end{abstract}

\section{Introduction}

\label{sec:intro}

In offline goal-conditioned reinforcement learning (GCRL) \citep{kaelbling1993, park2025ogbench}, the value function is often fit by temporal difference (TD) learning, which regresses each value toward a target built from the next state. The problem is that the target is itself an estimate, so each backup inherits the error of the last \citep{park2025horizon}. A goal $T$ steps away needs $O(T)$ backups. Divide and conquer removes this dependence on the horizon by composing two shorter estimates through a subgoal. So, the same goal now needs only $O(\log T)$ updates \citep{park2026trl}. This composition is possible because temporal distances obey the triangle inequality \citep{kaelbling1993, dhiman2018, wang2023qrl, piekos2023}.

Transitive RL (TRL) ~\citep{park2026trl} turns this triangle inequality into a practical update rule. However, the rule assumes deterministic dynamics, and the assumption sits at its base case. Under stochastic dynamics, the same action can lead to several successors, and a logged transition shows only one of them. The update rule chains these base cases together. So, it values the best case, where every random transition goes the agent's way. This overestimates the value ~\citep{myers2026offline, bao2026nftr}, and the gap does not shrink with more data. On top of this, the subgoal has to lie between the state and the goal within a trajectory. Without one, there is nothing to compose, and the state-goal pair never receives a target.



\begin{figure}[t]
\centering
\includegraphics[width=\textwidth]{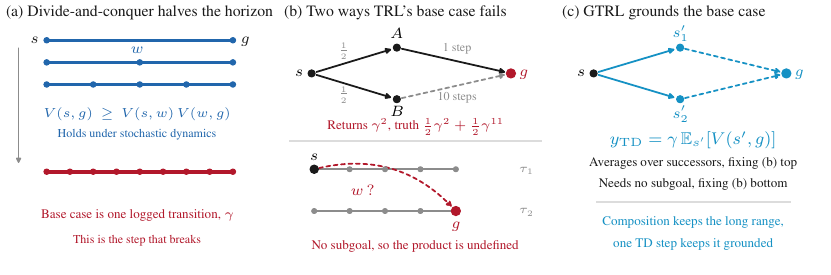}
\caption{\textbf{Where divide and conquer breaks.} (a) The transitive rule halves the horizon down to one logged transition, and base case is the step that breaks. (b) It breaks twice. Under stochastic dynamics, it reads the lucky successor as the agent's own choice (top), and it has no subgoal when the state and the goal lie on different trajectories (bottom). (c) A one-step TD target averages over the successors and needs no subgoal, so it repairs both, while the composition keeps the long horizon.}
\label{fig:concept}
\end{figure}

Our key insight is that both faults have the same fix, and that fix is TD learning, which divide and conquer originally set out to replace. TD learning is slow and faulty over a long horizon, but over a single step it is correct. Its target is an expectation over the next state, so the randomness is averaged. It also needs only a transition and a goal, so every state-goal pair gets a target. The triangle inequality still holds under stochastic dynamics \citep{myers2024}. So, we keep the value composition and ground it in a one-step target. Figure~\ref{fig:concept} illustrates the argument.


In this work, we introduce the resulting algorithm \textbf{Grounded Transitive RL (GTRL)}. It has two components, and the first is counterfactual goals. We sample goals from the whole dataset as well as from the state's own trajectory \citep{andrychowicz2017her}, and give every state-goal pair a one-step target. Where a subgoal also exists, the divide-and-conquer target and the TD target are both valid, so we take the larger.
The second is hindsight weighting. A goal from the state's own trajectory lies downstream of the successor that happened. So, under stochastic transitions, the critic credits the action with a goal only one successor could reach. We remove this bias by weighting each such goal with a ratio of state occupancies \citep{blier2021, schramm2023usher}. Every term in the ratio comes from the critic, so it costs one forward pass and no new network.



\textbf{Contributions.} Our contributions are a diagnosis of where divide and conquer breaks and an algorithm that repairs it. The transitive update converges to the value of the luckiest route through the data, and it never trains the state-goal pairs that no trajectory connects.
GTRL grounds that rule in a one-step expectation, and the whole change costs one extra forward pass. We evaluate our algorithm on OGBench's \citep{park2025ogbench} long-horizon stochastic, deterministic, and stitch tasks, where it achieves the highest average success rate without trading one setting for another.


\section{Related Work}
\label{sec:related}

\paragraph{Offline goal-conditioned RL.} Offline RL learns a policy from a fixed dataset of tasks, and one of its fundamental problems is that it overestimates the value of actions the dataset never shows \citep{fujimoto2019bcq,kumar2019bear,levine2020offline}. Existing solutions involve making the value function conservative \citep{kumar2020cql} or keeping the policy close to the behavior policy \citep{wu2019brac,fujimoto2021td3bc,tarasov2023rebrac,park2025fql}. 
Goal-conditioned RL asks for one more thing: a single policy must reach any state from any other, and its training goals usually come from hindsight relabeling \citep{kaelbling1993,andrychowicz2017her,park2025ogbench}. Many offline goal-conditioned methods adapt offline RL or supervised learning to goals \citep{ghosh2021gcsl,yang2022wgcsl,park2023hiql,ke2025cgcivl}. Others instead use a structure that only goal-reaching has, such as quasimetric value functions \citep{wang2023qrl}, hierarchical policies \citep{park2023hiql}, or a contrastive, probabilistic view of the value \citep{eysenbach2021clearning,eysenbach2022crl,zheng2024cpc}. Our proposed method belongs to this second group, and the structure it uses is the triangle inequality that goal-reaching values get from shortest paths.

\paragraph{Composition through subgoals.} In deterministic environments, the best route to a goal is at least as good as any route that must pass through a given intermediate state. So the optimal goal-conditioned value satisfies $V^*(s,g) \geq V^*(s,w)\,V^*(w,g)$ for every intermediate state $w$, an idea that goes back to the earliest work on goal-conditioned RL \citep{kaelbling1993}. Later methods have built this inequality into the architecture of the value function \citep{pitis2020inductive,wang2022iqe,liu2023mrn,wang2023qrl,myers2024}, or used it to plan through subgoals or suggest the subgoals altogether.  \citep{eysenbach2019search,nasiriany2019leap,parascandolo2020dcmcts,jurgenson2020,chanesane2021ris,li2021taskreduction}. We utilize the value composition property of this triangle inequality \citep{kaelbling1993,dhiman2018,jurgenson2020,piekos2023}. Joining two shorter estimates through a subgoal allows us to implement divide-and-conquer, which, in the best case, needs only logarithmic Bellman recursions in the horizon. Other methods such as $n$-step returns and chunked critics shorten the recursion only by a constant factor, and Monte Carlo targets remove it but add variance on top \citep{park2025horizon,li2025qchunking,li2026dqc,park2026trl}. Yet, without planning, value backup methods had mostly been demonstrated to work in two-dimensional mazes and grid worlds. Transitive RL \citep{park2026trl} scaled them up with two changes. It replaced the maximum over subgoals with an in-sample expectile \citep{kostrikov2022iql}, and it restricts subgoals to states on the same trajectory as the state and the goal. Both of these changes assume deterministic dynamics.

\paragraph{Stochastic dynamics.} When the dynamics are stochastic, prior work changes one of two things: the distance being estimated, or the relabeled data it is trained on. In terms of distance, a temporal distance built from contrastive successor features meets the triangle inequality even under stochastic dynamics \citep{myers2024}, while temporal metric distillation learns such distances offline with a quasimetric parameterization \citep{myers2026offline}. Distributional distance classifiers estimate the entire distribution of distances to the goal rather than one single value \citep{akella2023distributional}. In terms of data, the problem is that hindsight relabeling is unbiased only when the dynamics are deterministic. Otherwise, it overvalues the lucky outcomes \citep{plappert2018multigoal,blier2021unbiased}. Outcome-conditioned supervised methods fail for the same reason, unless they condition on quantities that the environment's randomness does not determine \citep{paster2022luck,yang2023doc}. USHER \citep{schramm2023usher} removes the relabeling bias with an importance weight on the Bellman error. However, its weight comes from a separately trained density of future goals, which is a successor representation \citep{dayan1993sr}, and is derived for a one-step backup in the online setting. 
None of these methods changes the divide-and-conquer transitive update \citep{park2026trl}. We fix the divide-and-conquer value composition with a one-step TD update, so the value stays grounded when the dynamics branch, and we take its relabeling correction from the critic, which only costs one forward pass and needs no extra network.

\section{Preliminaries}

\label{sec:prelim}

\paragraph{Problem setting.}
We consider a controlled Markov decision process $\mathcal{M} = (\mathcal{S}, \mathcal{A}, p)$ with state space $\mathcal{S}$, action space $\mathcal{A}$ and transition dynamics $p(s' \mid s, a)$. For offline goal-conditioned reinforcement learning (GCRL), we assume that we are given an unlabeled dataset $\mathcal{D} = \{\tau^{(i)}\}_i$ of trajectories $\tau = (s_0, a_0, s_1, \ldots, s_T)$ with no further access to the environment. In GCRL a task is defined by a state to reach rather than by a reward \citep{kaelbling1993, schaul2015uvfa}. Here, goals lie in state space $\mathcal{S}$, the rewaard is the sparse indicator ${0 [s \neq g], 1 [s = g]}$, and one policy function $\pi(a \mid s, g)$ serves a whole family of tasks. For our case, we use the hitting-time form of the objective \citep{wang2023qrl}, in which the goal is absorbing and the reward is collected upon arrival.

Let $T^{\pi}(s,g) = \min\{t \geq 0 : s_t = g\}$ be the arrival time of policy $\pi(\cdot \mid \cdot, g)$ started at $s_0 = s$, and $T^{\pi}(s,g) = \infty$ if it never arrives. The value function and its corresponding $Q$ function $Q(s, a, g) : \mathcal{S} \times \mathcal{A} \times \mathcal{S} \to \mathbb{R}$ are: 
\begin{equation}
    V^{\pi}(s,g) = \mathbb{E}\bigl[\gamma^{\,T^{\pi}(s,g)}\bigr],
    \qquad
    Q^{\pi}(s,a,g) = \mathbb{E}\bigl[\gamma^{\,T^{\pi}(s,g)} \mid a_0 = a\bigr],
    \label{eq:value}
\end{equation}
where $\gamma \in (0,1)$ is the discount factor and $\gamma^{\infty} = 0$. A larger value means a shorter expected trip. If the episode terminates at each step with probability $1 - \gamma$, then $V^{\pi}(s,g)$ is the probability of reaching $g$ before the termination. We define optimal $V^{*}$ and $Q^{*}$ as the maxima over policy $\pi$. The problem is to estimate $Q^{*}$ from offline dataset $\mathcal{D}$ and extract a policy from it.

\paragraph{Temporal distances.}
Traditionally in offline GCRL \citep{eysenbach2019search, janner2022planning,park2023hiql, wang2023qrl, park2026trl}, the dynamics are assumed to be deterministic unless stated otherwise. Therefore, arrival time is not random and the optimal policy takes the shortest route. The temporal distance $d^{*}(s,g)$ is the smallest number of steps from state $s$ to goal $g$. So, equation~\ref{eq:value} then collapses to $V^{*}(s,g) = \gamma^{d^{*}(s,g)}$. Optimal goal-conditioned values are a monotone reparameterization of a shortest-path distance \citep{kaelbling1993, wang2023qrl}, and it inherits the triangle inequality ~\citep{myers2026offline} that distances obey. This is the structure the value learning methods below exploit.

\subsection{Offline Goal-Conditioned Value Learning}
\label{sec:prelim-value}

\paragraph{Temporal difference learning.}

Since the task ends on arrival, Equation~\ref{eq:value} satisfies a one-step recursion: $Q^{*}(s,a,g)$ is $1$ when $s = g$, and $\gamma\,\mathbb{E}_{s' \sim p(\cdot \mid s,a)}[V^{*}(s',g)]$ otherwise, with $V^{*}(s,g) = \max_{a} Q^{*}(s,a,g)$. Temporal difference (TD) learning fits a critic to this recursion by regression \citep{mnih2013}, computing the target from a target network $\bar{Q}$. Each backup carries value information one step, so a goal $T$ steps away needs $O(T)$ value backups. Every one is a regression whose error the next inherits. This dependence on the horizon is one of the bottlenecks to scaling offline RL to long-horizon tasks \citep{park2025horizon}.

Fitting this recursion requires $\max_{a'} Q(s',a',g)$, which is unsafe offline because the maximum can pick an out of distribution action, and the critic has no data to correct its value \citep{levine2020offline}. Implicit Q-learning \citep{kostrikov2022iql, park2023hiql} approximates it by an asymmetric regression over in-sample actions instead. For a loss $L$, a prediction $x$, a target $y$ and a level $\kappa \in [0.5, 1)$,
\begin{equation}
    L_{\kappa}(x,y) = \left| \kappa - \mathbb{I}(x > y) \right| \cdot L(x,y)
    \label{eq:expectile}
\end{equation}
penalizes underestimating the target by $\kappa$ and overestimating it by $1 - \kappa$ \citep{newey1987}. Changing $\kappa$ changes what the critic learns. It fits the average of the targets at $\kappa = 0.5$ \citep{brandfonbrener2021} and moves closer to the largest target as $\kappa$ grows toward $1$.

\paragraph{Divide-and-conquer value learning.}
Divide and conquer replaces the one-step recursion with two composable value estimates. Let $w$ be any intermediate state, which we define as a subgoal. Going from $s$ to $g$ through $w$ is one of many available routes. So, it cannot be shorter than the optimal route. Distances therefore satisfy the triangle inequality $d^{*}(s,g) \leq d^{*}(s,w) + d^{*}(w,g)$ \citep{kaelbling1993, dhiman2018, wang2023qrl, piekos2023}. Since $V^{*}(s,g) = \gamma^{d^{*}(s,g)}$, the value decomposes into a product, i.e., $V^{*}(s,g) \geq V^{*}(s,w)\,V^{*}(w,g)$, and a subgoal on a shortest route makes it an equality. Taking the best subgoal therefore gives an update rule:
\begin{equation}
    Q(s,a,g) \;\longleftarrow\; \max_{w \in \mathcal{S},\, a_w \in \mathcal{A}} Q(s,a,w)\, Q(w,a_w,g),
    \label{eq:transitive}
\end{equation}
where $Q(s,a,g)$ is $1$ when $s = g$ and $\gamma$ when agent reaches goal $g$ in one step. Iterating this rule converges to $Q^{*}$ in the tabular case \citep{dhiman2018, piekos2023}. Each update joins two estimates instead of adding a step, so at best a horizon of $T$ needs $O(\log T)$ updates \citep{park2026trl} instead of $O(T)$.
The cost is the maximum over intermediate state $w$. It searches the whole state space, so a learned $Q$ picks whichever subgoal it overvalues, and the product inherits the error of both estimators \citep{levine2020offline}. 
This is why methods built on this rule have largely been shown on small domains \citep{dhiman2018, jurgenson2020, piekos2023}.


\paragraph{Policy extraction.}
Since $\arg\max_{a} Q(s,a,g)$ has no closed form, a critic does not by itself give a policy when the action space is continuous. We extract policy with reparameterized gradients \citep{fujimoto2021td3bc, park2024bottleneck}, which train policy $\pi_{\phi}$ to maximize the critic at its own sampled action under a behavior cloning term:
\begin{equation}
    J(\phi) = \mathbb{E}_{(s,g) \sim \mathcal{D},\; a^{\pi} \sim \pi_{\phi}(\cdot \mid s,g)}\bigl[Q(s,a^{\pi},g)\bigr]
    \;+\; \alpha_{\mathrm{BC}}\, \mathbb{E}_{(s,a,g) \sim \mathcal{D}}\bigl[\log \pi_{\phi}(a \mid s,g)\bigr],
    \label{eq:ddpgbc}
\end{equation}
where $\alpha_{\mathrm{BC}}$ sets how tightly the policy is tied to the data and can be tuned per environment.


\section{Methodology}

\label{sec:method}

\begin{figure}[t]
\centering
\includegraphics[width=\textwidth]{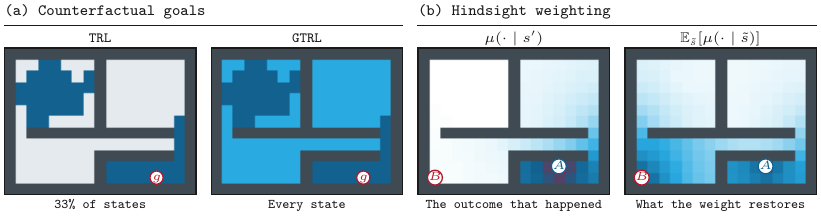}
\caption{\textbf{Method Overview.} We illustrate both components on a teleport maze, where one action leads to $A$ or to $B$ with probability $\tfrac{1}{2}$. Every panel is computed exactly, and each arrow runs from what TRL does to what GTRL does. (a) TRL trains only the states that reach $g$ on a shared trajectory, while GTRL trains all of them. (b) A hindsight goal is picked after the environment has chosen a successor, so it comes from that one outcome alone. The weight corrects it back to the average over every successor the action could have produced.}
\label{fig:method}
\end{figure}

In this work, we extend upon the prior work Transitive RL (TRL) \citep{park2026trl} and develop an effective value learning algorithm for offline GCRL. For deterministic environments, Transitive RL makes Equation~\ref{eq:transitive} practical in two ways. It replaces the maximum over subgoals with the asymmetric regression of Equation~\ref{eq:expectile} and it draws in-trajectory subgoals. So, for a trajectory $\tau \in \mathcal{D}$ and indices $i < k < j$, the critic loss becomes
\begin{equation}
    \mathcal{L}_{\mathrm{TRL}}(Q) = \mathbb{E}_{\substack{\tau \sim \mathcal{D},\; i < j \\ k \sim \mathrm{Unif}\{i, \ldots, j-1\}}}
    \Bigl[\, \rho(s_i,s_j)\; L_{0.7}\bigl(Q(s_i,a_i,s_j),\; \bar{Q}(s_i,a_i,s_k)\,\bar{Q}(s_k,a_k,s_j)\bigr) \Bigr],
    \label{eq:trlloss}
\end{equation}
where $\rho(s_i,s_j) = (1 + \log_{\gamma} \bar{Q}(s_i,a_i,s_j))^{-\lambda}$ puts more weight on short segments. We change two parts of this loss. The first, counterfactual goals, draws goals from outside the trajectory and gives them a one-step target. The second, hindsight weighting, corrects the hindsight bias caused by stochastic state transitions. The next sections explain both choices in detail.


\subsection{The Problem}
\label{sec:method-why}

The base case of Equation~\ref{eq:transitive} says that a transition in the dataset can be taken whenever the agent wants. Under stochastic dynamics it is not, because the agent chooses the action but the environment dynamics chooses the successor. 

What does this cost? Let $G$ be the directed graph on state space $\mathcal{S}$ with an edge from $s$ to $s'$ whenever $p(s' \mid s, a) > 0$ for some action $a$. Let $d_{G}(s,g)$ be the shortest path length from $s$ to $g$ in $G$. So, $d_{G}(s,g)$ is the fewest steps needed to reach goal $g$ from state $s$ if the randomness always goes the agent's way. For simplicity, we state the result in the state value form of Equation~\ref{eq:transitive}, $V(s,g) \leftarrow \max_{w \in \mathcal{S}} V(s,w)\,V(w,g)$, which has the same base cases.


\begin{proposition}[Fixed point of the transitive update]
\label{prop:fixed-point}
Assume every edge of graph $G$ appears in dataset $\mathcal{D}$, and let $V_n$ be the result of applying the update $n$ times, starting from the base cases and $0$ elsewhere. Let $V_{\mathrm{DC}}(s,g) = \gamma^{\,d_{G}(s,g)}$ be the divide-and-conquer value, and let $\operatorname{diam}(G)$ be the largest finite $d_{G}(s,g)$. Then
\begin{enumerate}[label=(\roman*),leftmargin=2.2em,itemsep=2pt,topsep=3pt]
    \item $V_n$ increases pointwise to $V_{\mathrm{DC}}$, and reaches $V_{\mathrm{DC}}$ after $\lceil \log_2 \operatorname{diam}(G) \rceil$ steps.
    \item $V_{\mathrm{DC}} \geq V^{*}$, with equality at $(s,g)$ if and only if some policy reaches $g$ from $s$ in exactly $d_{G}(s,g)$ steps almost surely.
    \item If $p$ is deterministic, $V_{\mathrm{DC}} = V^{*}$.
\end{enumerate}
\end{proposition}

The proof is in Appendix~\ref{app:proofs}. Intuitively, the update rule chains its base cases together, so what it converges to is the value of the shortest route the agent could conceivably take. Part (iii) recovers the guarantee the rule has in the deterministic case \citep{dhiman2018, piekos2023}. Part (ii) says that once the two come apart the error has one sign, so the rule overestimates. This gap belongs to the fixed point, and more data does not close it.

\begin{example}[Teleporter]
\label{ex:teleport}
Let $s$ have a single action that leads to state $A$ or to state $B$ with probability $1/2$ for each. One step reaches $g$ from $A$, while from $B$ the only route to $g$ is a corridor of length $10$ steps. Nothing returns to $s$. Then $d_{G}(s,g) = 2$, so $V_{\mathrm{DC}}(s,g) = \gamma^{2}$, while $V^{*}(s,g) = \tfrac{1}{2}\gamma^{2} + \tfrac{1}{2}\gamma^{11}$. At $\gamma = 0.99$ that is $V_{\mathrm{DC}} = 0.980$ against $V^{*} = 0.938$, or a reported distance of $2$ steps against a true $6.4$.
\end{example}

The composition step does not fail here. The inequality $V^{*}(s,g) \geq V^{*}(s,w)\,V^{*}(w,g)$ holds under stochastic dynamics as well \citep{myers2024}, so composing two correct values cannot produce an overestimate on its own. The base case fails, and it fails at exactly stochastic transitions. Its natural replacement is the one-step expectation of the temporal difference backup in Section~\ref{sec:prelim-value}.

The second problem is about coverage rather than bias. A subgoal has to lie in between $s$ and $g$ on one trajectory, and so $g$ must appear after $s$ on that trajectory in the first place.


\begin{proposition}[Which pairs receive a target]
\label{prop:reachability}
Let $\mathcal{R}$ be the set of pairs $(s_i, s_j)$ with $i \leq j$ that lie on a common trajectory $\tau \in \mathcal{D}$. Every target formed by Equation~\ref{eq:trlloss} is for a pair in $\mathcal{R}$. For $(s,g) \notin \mathcal{R}$ the loss never produces a target, so the value there stays at whatever it was initialized to.

\end{proposition}



The two propositions now point at the same missing piece. One says the base case needs a one-step expectation, and the other says the pairs off the trajectory need a target at all. A single one-step term supplies both.

\subsection{Counterfactual Goals}
\label{sec:method-cf}

In Example~\ref{ex:teleport}, the pair $(s,g)$ starts showing up on trajectories that went to $B$ as well.
Offline GCRL datasets carry no goals, and so, goals are assigned in hindsight \citep{andrychowicz2017her}. For a state $s_i$ on a trajectory $\tau$, the standard sampler \citep{park2023hiql, park2025ogbench} returns one of three things. It gives back $s_i$ itself, or a later state $s_{i+k}$ on $\tau$ with offset $k$ drawn from a $\mathrm{Geometric}(1-\gamma)$ distribution, or a uniformly drawn state from the entire dataset. Transitive RL never uses the third case, because a random goal from the dataset has no subgoal between it and $s_i$. So, the product in Equation~\ref{eq:trlloss} is undefined. However, we use it. Such goals are picked without looking at the trajectory, so the targets for the same input now come from every outcome of a stochastic transition. 


The rest of this section is what that choice enables. Let $(s,a,s')$ be a transition from $\mathcal{D}$ and let $a'$ be the action logged at $s'$. We define $m = 1$ as decomposable goals, which lie ahead of state $s$ on the same trajectory, and $m = 0$ otherwise. TRL trains on the decomposable goals, and their target is $y_{\mathrm{DC}} = \bar{Q}(s,a,w)\,\bar{Q}(w,a_w,g)$ from Equation~\ref{eq:trlloss}. Non-decomposable goals have no subgoal, so nothing composes. Nevertheless, they still have a one-step target,
\begin{equation}
    y_{\mathrm{TD}} \;=\;
    \begin{cases}
        1 & \text{if } g = s,\\[2pt]
        \gamma & \text{if } g = s',\\[2pt]
        \gamma\, \bar{Q}(s',a',g) & \text{otherwise.}
    \end{cases}
    \label{eq:td}
\end{equation}
Equation~\ref{eq:td} is Equation~\ref{eq:trlloss} at subgoal offset one, with the first factor at its base case $\gamma$ and the second at the subsequent logged action. On a decomposable goal the target varies with the subgoal draw. So, setting the expectile $\kappa$ above $0.5$ in Equation~\ref{eq:expectile} maximizes over decompositions. On a non-decomposable goal the target varies with the environment's transition, and there we want the average over outcomes. So,
\begin{equation}
    \kappa(s,g) \;=\; 0.7 \ \text{ if } m = 1,
    \qquad
    \kappa(s,g) \;=\; 0.5 \ \text{ if } m = 0.
    \label{eq:kappa}
\end{equation}


Since Equation~\ref{eq:td} is defined wherever a transition exists, decomposable goals now carry two targets. Equation~\ref{eq:transitive} is a maximum over subgoals and the one-step case is one of them, so we take the larger,
\begin{equation}
    y_{\mathrm{A}} \;=\;
    \begin{cases}
        \max\{y_{\mathrm{DC}},\, y_{\mathrm{TD}}\} & \text{if } m = 1,\\[2pt]
        y_{\mathrm{TD}} & \text{if } m = 0.
    \end{cases}
    \label{eq:anchored}
\end{equation}
On a non-decomposable goal the subgoal set is empty, so the maximum runs over the one-step case alone.

\subsection{Hindsight Weighting}
\label{sec:method-hw}

$Q(s,a,g)$ averages over every successor that action $a$ could produce. Hindsight relabeling \citep{andrychowicz2017her} breaks this, because it picks the goal after the successor is already known. In Example~\ref{ex:teleport}, a trajectory that went to $A$ only gives back goals near $A$, and one that went to $B$ only goals near $B$. So, for a given goal $g$, the critic mostly sees the transitions that got closer to it.

Let $\mu(\cdot \mid s)$ be the discounted state occupancy\footnote{\textbf{Discounted State Occupancy, $\mu(g \mid s)$:} If the episode terminates at each step with probability $1 - \gamma$, then $\mu(g \mid s)$ is the probability of being at goal $g$ when it terminates.} of the behavior policy from state $s$. The trajectory offset is drawn from $\mathrm{Geometric}(1-\gamma)$ distribution with the same $\gamma$ the critic uses. So, a decomposable goal is a sample from $\mu(\cdot \mid s')$. We want the average occupancy $\mu(\cdot \mid \tilde{s})$ over the successors $\tilde{s}$ that $a$ could produce. So, we weight each decomposable goal by the ratio of the two.
The identity $\mu(g \mid s) = V(s,g)\,\mu(g \mid g)$ \citep{blier2021} splits the occupancy into a value and a factor that depends on goal $g$ alone, and that factor cancels in the ratio,
\begin{equation}
    h(s,a,s',g)
    \;=\; \frac{\mathbb{E}_{\tilde{s} \sim p(\cdot \mid s,a)}\bigl[\mu(g \mid \tilde{s})\bigr]}{\mu(g \mid s')}
    \;=\; \frac{\mathbb{E}_{\tilde{s}}\bigl[V(\tilde{s},g)\bigr]}{V(s',g)}
    \;=\; \frac{Q(s,a,g)}{\gamma\, V(s',g)}.
    \label{eq:hw}
\end{equation}
The numerator is the average over outcomes and the denominator is the outcome that occurred. So, at a stochastic transition, a successor state that landed close to goal $g$ is weighted down, and one that landed far from it is weighted up. In the implementation the denominator is $\bar{Q}(s',a',g)$, which Equation~\ref{eq:td} already computes. So, hindsight weighting $h$ costs one extra forward pass and no new network.


Two cases need no correction. First, under deterministic dynamics $p(\cdot \mid s,a)$ is a point mass, so $Q(s,a,g) = \gamma V(s',g)$ and $h = 1$ everywhere. Second, a non-decomposable goal is sampled from the whole dataset. So, it is picked without knowing the successor and keeps $h = 1$. 
In practice, we clip $h$ to $[1/(1+c),\, 1+c]$ following \citep{schramm2023usher}, with $c = 1$, and normalize it over the decomposable goals in the batch. This leaves the scale of the loss unchanged.










\begin{algorithm}[t]
\caption{Grounded Transitive RL (GTRL)}
\label{alg:gtrl}
\begin{algorithmic}[1]
    \Require Dataset $\mathcal{D}$, discount $\gamma$, clip constant $c$, target update rate $\eta$
    \State Initialize critic $Q_{\theta}$, target critic $\bar{Q}_{\bar{\theta}} \leftarrow Q_{\theta}$, and policy $\pi_{\phi}$
    \For{each gradient step}
        \State Sample transitions $(s,a,s')$ from $\mathcal{D}$ with the action $a'$ logged at $s'$, and a goal $g$ for each
        \State Set $m = 1$ where $g$ lies ahead of $s$ on the same trajectory, and $m = 0$ otherwise
        \State Where $m = 1$, sample a subgoal $w$ uniformly between $s$ and $g$ with its logged action $a_w$
        \State $y_{\mathrm{DC}} \leftarrow \bar{Q}(s,a,w)\,\bar{Q}(w,a_w,g)$ \Comment{Equation~\ref{eq:trlloss}}
        \State $y_{\mathrm{TD}} \leftarrow \gamma\,\bar{Q}(s',a',g)$, or a base case where $g = s$ or $g = s'$ \Comment{Equation~\ref{eq:td}}
        \State $y_{\mathrm{A}} \leftarrow \max\{y_{\mathrm{DC}},\, y_{\mathrm{TD}}\}$ where $m = 1$, and $y_{\mathrm{TD}}$ elsewhere \Comment{Equation~\ref{eq:anchored}}
        \State $\kappa \leftarrow 0.7$ where $m = 1$, and $0.5$ elsewhere \Comment{Equation~\ref{eq:kappa}}
        \State $h \leftarrow \bar{Q}(s,a,g) \,/\, \gamma\bar{Q}(s',a',g)$ where $m = 1$, and $1$ elsewhere \Comment{Equation~\ref{eq:hw}}
        \State Clip $h$ to $[1/(1+c),\, 1+c]$, then normalize it to mean $1$ over the goals with $m = 1$
        \State $\theta \leftarrow \theta - \nabla_{\theta}\, \mathcal{L}_{\mathrm{GTRL}}(Q_{\theta})$ \Comment{Equation~\ref{eq:gtrlloss}}
        \State $\phi \leftarrow \phi + \nabla_{\phi}\, J(\phi)$ \Comment{Equation~\ref{eq:ddpgbc}}
        \State $\bar{\theta} \leftarrow \eta\,\theta + (1 - \eta)\,\bar{\theta}$
    \EndFor
\end{algorithmic}
\end{algorithm}


\subsection{The Full Objective}
\label{sec:method-loss}

GTRL samples a transition $(s,a,s')$ from dataset $\mathcal{D}$ with the action $a'$ logged at $s'$, and a goal $g$ from the sampler of Section~\ref{sec:method-cf}. It also samples a subgoal $w$ between $s$ and $g$ when the goal is decomposable. Ultimately, the critic loss becomes
\begin{equation}
    \mathcal{L}_{\mathrm{GTRL}}(Q) \;=\;
    \mathbb{E}\Bigl[\; h \;\cdot\; \rho \;\cdot\;
    L_{\kappa(s,g)}\bigl(Q(s,a,g),\; y_{\mathrm{A}}\bigr) \;\Bigr].
    \label{eq:gtrlloss}
\end{equation}
The target $y_{\mathrm{A}}$ is Equation~\ref{eq:anchored}, the expectile $\kappa(s,g)$ is Equation~\ref{eq:kappa}, the hindsight weight $h$ is Equation~\ref{eq:hw}, and $\rho$ is the distance weight of Equation~\ref{eq:trlloss}. The loss $\mathcal{L}_{\mathrm{GTRL}}$ addresses both problems introduced in Section~\ref{sec:method-why}. It fits a composition where the data supports one, and a one-step expectation everywhere else. For policy extraction, we use Equation~\ref{eq:ddpgbc} from Section~\ref{sec:prelim-value}. 
Pseudo-code Algorithm for GTRL stating the full value update is presented in Algorithm~\ref{alg:gtrl}.

\section{Experiments}
\label{sec:exp}

We evaluate GTRL on nineteen OGBench \citep{park2025ogbench} tasks, in three groups. Teleport tasks randomly send the agent to one of several destinations, so their transitions are stochastic and the base case of Section~\ref{sec:method-why} fails on them. The stitch tasks have only short trajectories, and the agent must join them together to reach a goal. So, most state-goal pairs never lie on a common trajectory, and the transitive loss never trains them. Section~\ref{sec:exp-faults} reports both groups. The remaining ten are standard deterministic navigation and manipulation tasks, and Section~\ref{sec:exp-standard} reports those. Section~\ref{sec:exp-ablation} then ablates each part of GTRL to see what it contributes. Each task uses its \textit{-oraclerep-v0} variant, and we report the success rate over its five evaluation goals, averaged over four seeds and the last three checkpoints, as in ~\citep{park2026trl}.


\paragraph{Baselines.}
Our baselines are prior offline GCRL algorithms from four families, listed below:
\begin{itemize}[leftmargin=1.4em, itemsep=1pt, topsep=3pt, parsep=0pt]
    \item \textbf{Behavioral cloning}: BC and FBC \citep{park2025ogbench}
    \item \textbf{Temporal difference}: IVL and IQL \citep{kostrikov2022iql, park2023hiql}, TD and TD-n \citep{sutton2018}
    \item \textbf{Monte Carlo}: CRL \citep{eysenbach2022crl} and MC \citep{tian2021, shah2021}
    \item \textbf{Triangle inequality}: QRL \citep{wang2023qrl}, TDP \citep{kaelbling1993, dhiman2018, jurgenson2020}, COE \citep{piekos2023} and TRL \citep{park2026trl}
\end{itemize}
BC, FBC, IVL and IQL are the goal-conditioned variants \citep{park2025ogbench}. We present our results with standard deviations, and the best performances are marked in \textcolor{bestred}{\textbf{RED}}. Implementation details, full results and hyperparameters are explained in Appendix~\ref{app:exp}.


\subsection{Does GTRL repair the faults of divide and conquer?}
\label{sec:exp-faults}

\paragraph{Tasks.}
We experiment on nine tasks from OGBench \citep{park2025ogbench}. The agent is a point mass, an 8-DoF quadruped, a 21-DoF humanoid, or the same quadruped dribbling a soccer ball. In the teleport maze, a teleporter randomly moves the agent to one of three destinations, and one of them is a dead end. In the stitch datasets, the maze trajectories are cut into four-cell pieces, and the task needs up to eight of them joined together to reach its goal.

\begin{table}[t]
\centering
\caption{\textbf{Evaluation on teleport and stitch tasks.} We report success rate (\%) and standard deviation over four seeds on OGBench \citep{park2025ogbench} tasks. Experiments used \textit{oracle-rep} variants. Best results are marked \textcolor{bestred}{\textbf{RED}}, and the second best are underlined.}
\label{tab:faults}
\footnotesize\ttfamily
\setlength{\tabcolsep}{3pt}
\resizebox{\textwidth}{!}{%
\begin{tabular}{llcccccccccc}
\toprule
Environment & Dataset & BC & FBC & IVL & IQL & TD & CRL & MC & QRL & TRL & \textcolor{bestred}{\textbf{GTRL}} \\
\midrule
pointmaze-teleport-navigate & \multirow{4}{*}{Teleport} & \scell{25}{3} & \scell{28}{2} & \ssec{35}{3} & \scell{29}{4} & \scell{24}{7} & \scell{24}{9} & \scell{29}{4} & \scell{4}{4}  & \scell{26}{5} & \sbest{37}{7} \\
pointmaze-teleport-stitch   &                           & \scell{31}{9} & \scell{34}{6} & \ssec{43}{2} & \scell{29}{2} & \scell{36}{2} & \scell{0}{0}  & \scell{26}{1} & \scell{12}{5} & \scell{40}{4} & \sbest{47}{2} \\
antmaze-teleport-navigate   &                           & \scell{26}{3} & \scell{29}{3} & \scell{40}{4} & \scell{29}{3} & \scell{27}{3} & \sbest{50}{2} & \scell{27}{2} & \scell{28}{4} & \scell{32}{2} & \ssec{49}{2} \\
antmaze-teleport-stitch     &                           & \scell{31}{6} & \scell{31}{4} & \ssec{34}{3} & \scell{18}{4} & \scell{24}{2} & \scell{12}{4} & \scell{22}{0} & \scell{13}{4} & \scell{32}{4} & \sbest{36}{2} \\
\cmidrule(lr){1-12}
Average & & 28.2 & 30.5 & \underline{38.0} & 26.2 & 27.7 & 21.5 & 26.0 & 14.2 & 32.5 & \textcolor{bestred}{\textbf{42.2}} \\
\midrule
pointmaze-large-stitch     & \multirow{5}{*}{Stitch} & \scell{7}{5}  & \scell{20}{13} & \scell{12}{6} & \ssec{34}{2} & \scell{0}{0}  & \scell{0}{0}  & \scell{0}{0}  & \sbest{84}{15} & \scell{0}{0}  & \scell{19}{2} \\
antmaze-large-stitch       &                         & \scell{3}{3}  & \scell{6}{4}   & \ssec{18}{2}  & \scell{8}{1}  & \scell{4}{1}  & \scell{8}{5}  & \scell{5}{1}  & \ssec{18}{2}   & \scell{8}{1}  & \sbest{31}{4} \\
humanoidmaze-medium-stitch &                         & \scell{29}{5} & \scell{38}{3}  & \scell{12}{2} & \scell{29}{7} & \scell{38}{1} & \ssec{59}{1} & \scell{35}{3} & \scell{18}{2}   & \scell{36}{0} & \sbest{70}{3} \\
humanoidmaze-large-stitch  &                         & \scell{6}{3}  & \scell{6}{1}   & \scell{1}{1}  & \scell{1}{1}  & \scell{3}{1}  & \ssec{12}{2} & \scell{7}{1}  & \scell{3}{1}    & \scell{5}{1}  & \sbest{18}{2} \\
antsoccer-arena-stitch     &                         & \scell{24}{8} & \scell{2}{3}   & \scell{21}{3} & \ssec{34}{2} & \scell{1}{1}  & \scell{2}{1}  & \scell{1}{0}  & \scell{1}{1}    & \scell{1}{0}  & \sbest{38}{2} \\
\cmidrule(lr){1-12}
Average & & 13.8 & 14.2 & 12.8 & 21.3 & 9.2 & 16.4 & 9.4 & \underline{24.8} & 9.8 & \textcolor{bestred}{\textbf{35.2}} \\
\bottomrule
\end{tabular}}
\end{table}

The top block of Table~\ref{tab:faults} reports the teleport tasks. GTRL averages 42.2\% and is ahead of the second-best GC-IVL (IVL) by 4.2 points (\textbf{+11\% relative}). Against the divide-and-conquer method TRL, GTRL is ahead on every task by 9.7 points on average. However, the one exception is antmaze-teleport-navigate, where the Monte Carlo method CRL \citep{eysenbach2022crl} finishes a point ahead of GTRL. The bottom block of Table~\ref{tab:faults} reports the stitch tasks. GTRL averages 35.2\% and is ahead of the second-best QRL by 10.4 points (\textbf{+42\% relative}). Against TRL, the margin is 25.4 points, and TRL itself finishes next to last at 9.8\%. The one exception is pointmaze-large-stitch, where the quasimetric method QRL \citep{wang2023qrl} reaches 84\%, and nothing else passes 34\%. 


\subsection{Does GTRL help on standard tasks as well?}
\label{sec:exp-standard}

\begin{table}[t]
\centering
\caption{\textbf{Evaluation on standard deterministic tasks.} We report success rate (\%) and standard deviation over four seeds on ten deterministic OGBench \citep{park2025ogbench} tasks.}
\label{tab:standard}
\footnotesize\ttfamily
\setlength{\tabcolsep}{2pt}
\resizebox{\textwidth}{!}{%
\begin{tabular}{lcccccccccccc}
\toprule
Environment & BC & FBC & IVL & IQL & TD-n & CRL & MC & QRL & TDP & COE & TRL & \textcolor{bestred}{\textbf{GTRL}} \\
\midrule
pointmaze-large-navigate     & \scell{25}{3} & \ssec{71}{4} & \scell{48}{10} & \scell{34}{3} & \scell{31}{7} & \scell{33}{7} & \scell{4}{4}   & \scell{7}{7}  & \scell{30}{5} & \scell{34}{7} & \scell{33}{5} & \sbest{79}{6} \\
antmaze-large-navigate       & \scell{22}{4} & \scell{22}{4} & \scell{21}{6}  & \scell{37}{8} & \scell{55}{6} & \ssec{85}{7} & \scell{39}{5}  & \scell{67}{8} & \scell{27}{3} & \scell{26}{3} & \scell{46}{5} & \sbest{87}{1} \\
humanoidmaze-medium-navigate & \scell{7}{2}  & \scell{9}{3}  & \scell{24}{4}  & \scell{36}{3} & \scell{60}{3} & \ssec{72}{5} & \scell{49}{3}  & \scell{18}{17} & \scell{7}{0} & \scell{9}{2}  & \scell{57}{1} & \sbest{75}{4} \\
humanoidmaze-large-navigate  & \scell{2}{1}  & \scell{1}{1}  & \scell{3}{1}   & \scell{5}{1}  & \scell{20}{2} & \ssec{28}{5} & \scell{6}{2}   & \scell{3}{2}  & \scell{1}{1}  & \scell{2}{0}  & \scell{8}{1}  & \sbest{41}{9} \\
antsoccer-arena-navigate     & \scell{3}{1}  & \scell{17}{2} & \scell{62}{2}  & \sbest{77}{3} & \scell{64}{4} & \scell{35}{5} & \scell{42}{3}  & \scell{13}{1} & \scell{5}{1}  & \scell{5}{2}  & \scell{73}{4} & \ssec{74}{1} \\
cube-single-play             & \scell{7}{2}  & \scell{18}{5} & \scell{88}{2}  & \ssec{95}{1}  & \ssec{95}{2}  & \scell{63}{5} & \sbest{98}{1}  & \scell{6}{3}  & \scell{3}{1}  & \scell{13}{8} & \ssec{95}{2}  & \scell{93}{4} \\
cube-double-play             & \scell{1}{1}  & \scell{4}{2}  & \ssec{59}{2}   & \sbest{64}{4} & \scell{10}{2} & \scell{35}{3} & \scell{8}{2}   & \scell{1}{1}  & \scell{1}{0}  & \scell{0}{0}  & \scell{30}{5} & \scell{37}{1} \\
scene-play                   & \scell{4}{2}  & \scell{23}{2} & \scell{68}{10} & \scell{61}{3} & \scell{75}{3} & \scell{26}{5} & \scell{25}{5}  & \scell{6}{2}  & \scell{12}{1} & \scell{8}{2}  & \ssec{77}{2} & \sbest{84}{4} \\
puzzle-3x3-play              & \scell{1}{1}  & \scell{3}{1}  & \scell{2}{1}   & \scell{98}{0} & \ssec{99}{0}  & \scell{5}{1}  & \scell{81}{12} & \scell{1}{1}  & \scell{2}{0}  & \scell{2}{0}  & \ssec{99}{0} & \sbest{100}{0} \\
puzzle-4x4-play              & \scell{0}{0}  & \scell{1}{0}  & \scell{5}{2}   & \scell{28}{4} & \scell{6}{3}  & \scell{0}{0}  & \scell{5}{2}   & \scell{0}{0}  & \scell{0}{0}  & \scell{0}{0}  & \ssec{34}{4} & \sbest{57}{8} \\
\cmidrule(lr){1-13}
Average & 7.2 & 16.9 & 38.0 & 53.5 & 51.5 & 38.2 & 35.7 & 12.2 & 8.8 & 9.9 & \underline{55.2} & \textcolor{bestred}{\textbf{72.7}} \\
\bottomrule
\end{tabular}}
\end{table}

\begin{figure}[t]
\centering
\includegraphics[width=\textwidth]{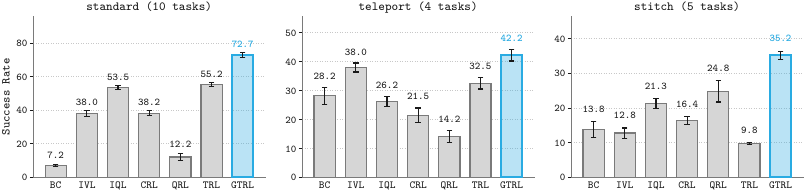}
\caption{\textbf{Average success rate in each environment.} Bars are the average over the tasks of that environment, and the error bars propagate the per-task seed standard deviations.}
\label{fig:summary-blue}
\end{figure}

We further evaluate our algorithm on ten standard deterministic tasks, spanning maze navigation, ball control, object manipulation, and puzzle solving \citep{park2025ogbench}. Our goal here is to determine whether grounding divide-and-conquer value can give comparable performance to existing GCRL algorithms on deterministic tasks as well. According to Table~\ref{tab:standard}, GTRL averages 72.7\% and is ahead of the second-best TRL by 17.5 points (\textbf{+32\% relative}). The largest gains are on the four maze navigation tasks, where GTRL adds between 18 and 46 points over the baselines.
The exceptions are \textit{antsoccer-arena-navigate} and the \textit{two cube} tasks, where IQL and MC methods finish ahead. Of these, every triangle inequality method struggles on \textit{cube-double-play}, and GTRL at 37\% is still the best of them.
GTRL therefore has the highest average in all three environment types, without trading one for another.
Figure~\ref{fig:summary-blue} shows performance on all three environments.




\subsection{Ablation}
\label{sec:exp-ablation}
We run ablations on three core components of GTRL (counterfactual goals, target selection, and hindsight weighting) and present our findings below:

\begin{wrapfigure}[12]{r}{0.48\textwidth}
\vspace{-\intextsep}
\centering
\includegraphics[width=0.46\textwidth]{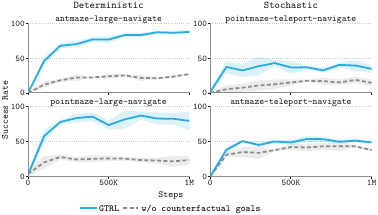}
\caption{Ablation study on \textbf{counterfactual goals.}}
\label{fig:abl-counterfactual}
\vspace{-\intextsep}
\end{wrapfigure}
\paragraph{Counterfactual goals.}
\label{sec:abl-counterfactual}
GTRL draws half of the critic batch from the whole dataset instead of the state's own trajectory, and gives each of those goals the one-step target of Equation~\ref{eq:td}. We ablate this by sampling in-trajectory goals alone, as in \citet{park2026trl}. This leaves the divide-and-conquer product as the only target. Figure~\ref{fig:abl-counterfactual} shows a sharp drop on every task, by up to 60.6 points on the deterministic tasks and 20.1 points on the stochastic ones. Counterfactual goals are what make the critic generalize past its own trajectories instead of memorizing them.




\newpage
\begin{wrapfigure}[11]{r}{0.48\textwidth}
\vspace{-\intextsep}
\centering
\includegraphics[width=0.46\textwidth]{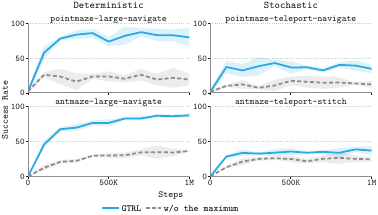}
\caption{Ablation study on \textbf{target selection.}}
\label{fig:abl-max}
\vspace{-\intextsep}
\end{wrapfigure}
\paragraph{Target Selection}
\label{sec:abl-max}
On decomposable goals, the divide-and-conquer product and the one-step target are both valid, and Equation~\ref{eq:anchored} takes the larger of the two. We ablate this by always taking the divide-and-conquer target on decomposable goals, and the one-step target on non-decomposable goals. Figure~\ref{fig:abl-max} shows a massive drop on the deterministic and the stochastic tasks, which is almost as costly as dropping the goals themselves. Target selection is what lets the one-step target fix the composition instead of just filling in for it.








\begin{wrapfigure}[12]{r}{0.48\textwidth}
\vspace{-\intextsep}
\centering
\includegraphics[width=0.46\textwidth]{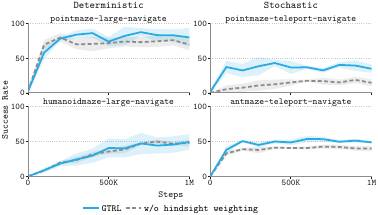}
\caption{Ablation study on \textbf{hindsight weighting.}}
\label{fig:abl-hindsight}
\vspace{-\intextsep}
\end{wrapfigure}
\paragraph{Hindsight weighting}
\label{sec:abl-hindsight}
Equation~\ref{eq:hw} weights each decomposable goal by how reachable it was from the other successors of the same action. We ablate it by setting $h = 1$ everywhere, which returns the loss to plain hindsight relabeling. Figure~\ref{fig:abl-hindsight} shows a small drop in deterministic tasks as the transition probability is one. However, on stochastic tasks, enabling hindsight weighting leads to improved performance. 
Further experiments on the choice of expectile and the goal sampling probability are reported in Appendix~\ref{app:exp}.

\section{Conclusion}
\label{sec:conclusion}

We presented GTRL, an offline goal-conditioned value learning algorithm that grounds divide-and-conquer updates in temporal differences. Divide-and-conquer reaches a distant goal in $O(\log T)$ updates rather than $O(T)$, but under stochastic dynamics the rule converges to the value of the luckiest route through the data. It also never trains the state-goal pairs that no trajectory connects. GTRL removes both faults with a one-step TD target, which averages over successors and needs no subgoal. Divide-and-conquer value composition remains unchanged, so the long horizon survives. Across nineteen OGBench tasks, GTRL has the highest average success rate on stochastic, stitch, and deterministic tasks, and it does not trade performance in one environment for another.


\textbf{Limitations.} Divide-and-conquer value composition still draws its subgoal from the data, so unconnected pairs rely on one-step backups. The hindsight weight is also read off the critic, so it inherits the critic's error. Natural next steps are to learn subgoals for unconnected pairs and to estimate the weight independently of the critic.



\subsection*{AI use statement}

We used Claude Opus, a frontier model, to edit the prose of this paper. Its role was to tighten wording and to shorten passages that ran long. The research question, the method, the proofs, the code, and the experiments are our own, and no part of them was produced by a generative AI tool. We read every edit Claude suggested, and we take responsibility for the final content of this paper.

\subsection*{Ethics statement}

This work studies value learning on simulated offline benchmarks. Every dataset we use comes from OGBench \citep{park2025ogbench}, which is public and released for research use, and none of it involves human subjects or personally identifying information. Our contribution is a value learning algorithm for goal reaching, so it carries the same risks as offline reinforcement learning in general and none that are specific to it. We declare no conflicts of interest.

\subsection*{Reproducibility statement}

Our code is included in the supplementary material, and its README gives the commands needed to reproduce every number we report. We will release it publicly upon publication. Appendix~\ref{app:exp} describes the compute, the baselines, the datasets, the implementation and the evaluation protocol, and Appendix~\ref{app:exp-hparams} lists every hyperparameter. Algorithm~\ref{alg:gtrl} states the update in full. The assumptions behind each of our claims are stated where the claim is made, and the proofs are in Appendix~\ref{app:proofs}.

\bibliographystyle{iclr2027_conference}
\bibliography{iclr2027_conference}
\newpage
\appendix
\label{sec:appendix}

\section{Proofs}
\label{app:proofs}

In this section, we provide the grounds for the two propositions of Section~\ref{sec:method-why}. Let $d = d_{G}$ be the shortest path length in the graph $G$ defined there, with $d(s,g) = \infty$ when $g$ is unreachable from $s$. Let $\mathcal{T}$ be the operator that applies the update at every pair at once, so $V_{n+1} = \mathcal{T}V_n$. On a base pair $\mathcal{T}$ writes the base value, which is $1$ at $s = g$ and $\gamma$ when $(s,g)$ is an edge of $G$, and at every other pair it writes $\max_{w \in \mathcal{S}} V(s,w)\,V(w,g)$.

\subsection{Proof of Proposition~\ref{prop:fixed-point}}
\label{app:proof-fixed-point}

\paragraph{Part (i).}
$\mathcal{T}$ is monotone. All values lie in $[0,1]$, so $V \leq V'$ gives $V(s,w)V(w,g) \leq V'(s,w)V'(w,g)$ at every subgoal $w$, and the base pairs do not depend on $V$ at all. Also $\mathcal{T}V \geq V_0$ for every $V$, since $\mathcal{T}V$ equals $V_0$ on the base pairs and is non-negative elsewhere. So $V_1 \geq V_0$, and monotonicity gives $V_{n+1} \geq V_n$ for every $n$.

Divide-and-conquer value $V_{\mathrm{DC}}$ is a fixed point of $\mathcal{T}$. Joining a shortest path from state $s$ to subgoal $w$ with a shortest path from $w$ to $g$ gives a walk from $s$ to $g$, so $d(s,g) \leq d(s,w) + d(w,g)$. Since discount factor $\gamma < 1$ this gives $V_{\mathrm{DC}}(s,w)\,V_{\mathrm{DC}}(w,g) \leq V_{\mathrm{DC}}(s,g)$. The choice $w = s$ attains it, because $d(s,s) = 0$. On a base pair $V_{\mathrm{DC}}$ already takes the base value, since $d(s,s) = 0$ and $d(s,g) = 1$ on an edge. So $\mathcal{T}V_{\mathrm{DC}} = V_{\mathrm{DC}}$, and as $V_0 \leq V_{\mathrm{DC}}$, monotonicity gives $V_n \leq V_{\mathrm{DC}}$ for every $n$.

It remains to show that $V_n(s,g) = V_{\mathrm{DC}}(s,g)$ whenever $d(s,g) \leq 2^{n}$. At $n = 0$ these are exactly the base pairs. Assume it at $n$ and take a pair with $2^{n} < m := d(s,g) \leq 2^{n+1}$. Then $m \geq 2$, so $(s,g)$ is not a base pair and $\mathcal{T}$ takes the maximum there. Fix a shortest path $s = x_0, \ldots, x_m = g$ and let $w = x_{\lceil m/2 \rceil}$. Every prefix and suffix of a shortest path is itself shortest, so $d(s,w) = \lceil m/2 \rceil$ and $d(w,g) = \lfloor m/2 \rfloor$, and both are at most $2^{n}$. The inductive hypothesis applies to each leg, and $w$ is one of the candidates, so
\[
    V_{n+1}(s,g) \;\geq\; V_n(s,w)\,V_n(w,g) \;=\; \gamma^{\lceil m/2 \rceil + \lfloor m/2 \rfloor} \;=\; \gamma^{m} \;=\; V_{\mathrm{DC}}(s,g),
\]
and $V_{n+1} \leq V_{\mathrm{DC}}$ makes it an equality. Pairs that already agree stay in agreement, since $V_n$ is nondecreasing and bounded above by $V_{\mathrm{DC}}$.

If $d(s,g) = \infty$ then $V_{\mathrm{DC}}(s,g) = 0$. Every $w$ then has $d(s,w) = \infty$ or $d(w,g) = \infty$, since two finite legs would join into a finite walk. So $V_n \leq V_{\mathrm{DC}}$ forces one factor of every product to be zero, and $V_n(s,g) = 0$ for every $n$.

Let the graph diameter of the transition graph $G$ be $\operatorname{diam}(G)$, which is restricted to reachable pairs. Now take $n = \lceil \log_2 \operatorname{diam}(G) \rceil$, the least $n$ with $2^{n} \geq \operatorname{diam}(G)$. Every pair with finite $d$ satisfies $d(s,g) \leq \operatorname{diam}(G) \leq 2^{n}$ and so agrees with $V_{\mathrm{DC}}$, and the remaining pairs agree by the previous paragraph. Hence $V_n = V_{\mathrm{DC}}$.

\paragraph{Part (ii).}
Fix a policy $\pi$ and a realization that reaches $g$ from $s$. Every transition it takes has positive probability under $p$, so it is an edge of $G$, and the realization traces a walk from $s$ to $g$ of length $T^{\pi}(s,g)$. A walk is no shorter than the shortest path, so $T^{\pi}(s,g) \geq d(s,g)$. A realization that never reaches $g$ has $T^{\pi}(s,g) = \infty$, so the bound holds there too. Therefore $\gamma^{T^{\pi}(s,g)} \leq \gamma^{d(s,g)}$ almost surely, and taking the expectation and then the maximum over $\pi$ gives $V^{*} \leq V_{\mathrm{DC}}$.

A random variable bounded above by a constant has that constant as its mean only if it equals it almost surely. So $\mathbb{E}[\gamma^{T^{\pi}(s,g)}] = \gamma^{d(s,g)}$ holds exactly when $T^{\pi}(s,g) = d(s,g)$ almost surely. Section~\ref{sec:prelim} defines $V^{*}$ as a maximum over $\pi$, so equality at $(s,g)$ means some policy meets that condition. When $d(s,g) = \infty$ no policy reaches $g$ and both sides are $0$.

\paragraph{Part (iii).}
Let $p$ be deterministic and let $(s,s')$ be an edge of $G$. Then $p(s' \mid s,a) > 0$ for some $a$, and a deterministic $p(\cdot \mid s,a)$ puts all of its mass on one successor, so $p(s' \mid s,a) = 1$. The agent can therefore traverse any edge of $G$ at will. A shortest path visits distinct states, so the policy that takes the action leading from $x_t$ to $x_{t+1}$ at each of them is well defined, and it reaches $g$ in exactly $d(s,g)$ steps. Part (ii) then gives $V_{\mathrm{DC}}(s,g) = V^{*}(s,g)$. If $d(s,g) = \infty$ both sides are $0$. \hfill$\square$

Part (ii) also locates where the rule leaves $V^{*}$. Let $(s,g)$ be an edge of $G$ with $s \neq g$. Every policy has $T^{\pi}(s,g) \geq 1$ almost surely, so $V^{*}(s,g) \leq \gamma$, which is the base value, with equality exactly when $p(g \mid s,a) = 1$ for some $a$. At $s = g$ the base value is $1 = V^{*}(s,s)$. So the base case sits above $V^{*}$ at exactly the transitions whose outcome the agent cannot force, which is the claim made in Section~\ref{sec:method-why}.

\subsection{Proof of Proposition~\ref{prop:reachability}}
\label{app:proof-reachability}

Equation~\ref{eq:trlloss} draws one trajectory $\tau \in \mathcal{D}$, two indices $i < j$ on it, and a subgoal index $k \in \{i, \ldots, j-1\}$. The prediction it regresses is $Q(s_i,a_i,s_j)$, so the pair that draw forms a target for is $(s_i,s_j)$. Both states lie on $\tau$ and $i < j$, so $(s_i,s_j) \in \mathcal{R}$. No draw forms a target for any other pair, so every target of the loss is for a pair in $\mathcal{R}$. The two factors of that target are read at $(s_i,s_k)$ and $(s_k,s_j)$, which lie on $\tau$ in that order as well, so the loss touches the critic only at pairs in $\mathcal{R}$.

Now take $(s,g) \notin \mathcal{R}$. No draw regresses $Q(s,a,g)$, so the loss does not depend on the value at that pair. In the tabular case its gradient with respect to that entry is zero at every step, and the entry keeps the value it was initialized to. \hfill$\square$

With a shared function approximator the entry does move, but only because the network generalizes from the pairs in $\mathcal{R}$. The loss itself still says nothing about it.

\subsection{Where the GTRL Fixed Point Lies}
\label{app:bracket}

The two propositions above are about the transitive update. This one bounds what GTRL converges to. Let $\beta$ be the policy that collected $\mathcal{D}$ and let $Q^{\beta}(s,a,g) = \mathbb{E}[\gamma^{\,T^{\beta}(s,g)} \mid a_0 = a]$ be its hitting time value. Let $\mathcal{T}_{\mathrm{A}}$ be the operator that maps a table $Q$ to the fit of Equation~\ref{eq:gtrlloss} against the targets $Q$ produces.

We work in the same tabular setting as before, with $\mathcal{S}$ and $\mathcal{A}$ finite and values in $[0,1]$, and we assume three things. Each entry is fit exactly, so it equals the $\kappa(s,g)$ expectile of its own targets. The successor $s'$ in Equation~\ref{eq:td} is drawn from $p(\cdot \mid s,a)$ and the next action is the logged $a'$, which is what the weight of Equation~\ref{eq:hw} restores. And $p > 0$, so every pair of dataset states is drawn with positive probability.

\begin{proposition}[Where the GTRL fixed point lies]
\label{prop:bracket}
Iterating $\mathcal{T}_{\mathrm{A}}$ from $Q_0 = 0$ converges to a fixed point $Q_{\mathrm{A}}$. At every state-action pair of $\mathcal{D}$ and every goal $g$ in $\mathcal{D}$,
\begin{equation*}
    Q^{\beta}(s,a,g) \;\leq\; Q_{\mathrm{A}}(s,a,g) \;\leq\; V_{\mathrm{DC}}(s,g),
    \qquad
    Q^{\beta}(s,a,g) \;\leq\; Q^{*}(s,a,g) \;\leq\; V_{\mathrm{DC}}(s,g).
\end{equation*}
\end{proposition}

The proof uses three properties of the expectile of Equation~\ref{eq:expectile} \citep{newey1987}. It lies between the smallest and the largest target. It does not fall when the targets rise or when $\kappa$ rises. At $\kappa = 0.5$ it is the mean of the targets, which is the statement in Section~\ref{sec:prelim-value}.

\paragraph{Monotonicity.}
The targets do not fall when $Q$ rises. Equation~\ref{eq:trlloss} multiplies two entries, both non-negative, Equation~\ref{eq:td} scales one entry by $\gamma$, and Equation~\ref{eq:anchored} takes their maximum. So $Q \leq Q'$ gives $\mathcal{T}_{\mathrm{A}}Q \leq \mathcal{T}_{\mathrm{A}}Q'$.

\paragraph{Upper bound.}
Suppose $Q(s,a,g) \leq V_{\mathrm{DC}}(s,g)$ at every entry. On a decomposable goal the target is $Q(s,a,w)\,Q(w,a_w,g) \leq \gamma^{\,d(s,w) + d(w,g)}$, and $d(s,g) \leq d(s,w) + d(w,g)$, so it is at most $V_{\mathrm{DC}}(s,g)$. The one-step target is $\gamma\,Q(s',a',g) \leq \gamma^{\,1 + d(s',g)}$, and $(s,s')$ is an edge of $G$, so $d(s,g) \leq 1 + d(s',g)$ and the same bound holds. Its two base cases are $1 = V_{\mathrm{DC}}(s,s)$ and $\gamma = V_{\mathrm{DC}}(s,s')$. Every target is therefore at most $V_{\mathrm{DC}}(s,g)$, and so is the fit. Since $Q_0 = 0 \leq V_{\mathrm{DC}}$, every iterate keeps the bound.

\paragraph{Lower bound.}
Let $\mathcal{T}_{\beta}$ be the one-step backup of $\beta$, that is $(\mathcal{T}_{\beta}Q)(s,a,g) = \gamma\,\mathbb{E}[Q(s',a',g)]$ for $s \neq g$, with $Q(g,\cdot,g) = 1$. It is a $\gamma$ contraction and its fixed point is $Q^{\beta}$. Equation~\ref{eq:anchored} is at least the one-step target at every goal, and Equation~\ref{eq:kappa} keeps $\kappa \geq 0.5$ at every goal, so the fit is at least the mean of the one-step target. Hence $\mathcal{T}_{\mathrm{A}}Q \geq \mathcal{T}_{\beta}Q$ for every $Q \geq 0$. Both operators are monotone and both start from $Q_0 = 0$, so $\mathcal{T}_{\mathrm{A}}^{\,n}0 \geq \mathcal{T}_{\beta}^{\,n}0$ at every $n$, and the right side converges to $Q^{\beta}$.

\paragraph{Convergence.}
$\mathcal{T}_{\mathrm{A}}0 \geq 0$, so monotonicity makes the iterates nondecreasing, and the upper bound keeps them below $V_{\mathrm{DC}}$. A nondecreasing bounded sequence converges, and $\mathcal{T}_{\mathrm{A}}$ is continuous on a finite table, so the limit $Q_{\mathrm{A}}$ is a fixed point and carries both bounds.

\paragraph{The second chain.}
$\beta$ is one policy, so $Q^{\beta} \leq Q^{*}$. For $s \neq g$ we have $Q^{*}(s,a,g) = \gamma\,\mathbb{E}_{s'}[V^{*}(s',g)]$, and Proposition~\ref{prop:fixed-point} gives $V^{*}(s',g) \leq \gamma^{\,d(s',g)} \leq \gamma^{\,d(s,g) - 1}$, so $Q^{*}(s,a,g) \leq V_{\mathrm{DC}}(s,g)$. At $s = g$ both sides are $1$. \hfill$\square$

The upper bound says what the maximum of Equation~\ref{eq:anchored} costs. Taking the larger of the two targets does not push the fixed point past the divide-and-conquer value of Proposition~\ref{prop:fixed-point}, so GTRL is no more optimistic than the rule it grounds.

The lower bound is what counterfactual goals buy. Outside $\mathcal{R}$ the transitive loss forms no target, so by Proposition~\ref{prop:reachability} the value there is whatever it was initialized to. GTRL has the one-step target there instead, and the bound floors it at $Q^{\beta}$, which is positive whenever $\beta$ reaches $g$ from $s$ with positive probability.


\section{Experimental Details}
\label{app:exp}


\subsection{Experimental Setup}
\label{app:exp-compute}

We ran every experiment on a single workstation with one NVIDIA GeForce RTX 5090 with 32\,GB of memory, an Intel Core i9-14900K with 24 cores, and 61\,GB of system memory, under Ubuntu 26.04 with CUDA 13.2. GTRL and all baselines are written in JAX 0.10.2, with Flax 0.12.8 for the networks and Optax 0.2.8 for the optimizer, on Python 3.11. The tasks come from OGBench \citep{park2025ogbench}, which runs on MuJoCo 3.12.0 through Gymnasium 1.3.0.

\subsection{Baselines}
\label{app:exp-baselines}

Offline GCRL algorithms differ mostly in how they fit the value function, so we group the baselines the same way as Section~\ref{sec:exp}. We use the reference implementations released with OGBench \citep{park2025ogbench} and Transitive RL \citep{park2026trl}.

\paragraph{Behavioral cloning.}
BC fits a goal-conditioned Gaussian policy to the dataset actions and learns no value at all. FBC replaces that policy with a flow matching model, which can fit an action distribution with several modes \citep{park2025ogbench}.

\paragraph{Temporal difference.}
IQL \citep{kostrikov2022iql} fits the critic with the one-step recursion of Section~\ref{sec:prelim-value} and handles the maximum over actions with the asymmetric regression of Equation~\ref{eq:expectile}. IVL is its goal-conditioned state value form \citep{park2023hiql}. TD applies the same backup over a single step, and TD-n over $n$ steps, clipping $n$ when the goal arrives sooner \citep{sutton2018}. All four carry the horizon dependence of Section~\ref{sec:prelim-value}.

\paragraph{Monte Carlo.}
CRL \citep{eysenbach2022crl} learns the value by telling the states a trajectory reached apart from states drawn at random. MC regresses the critic onto $\gamma^{T}$ read off the trajectory \citep{tian2021, shah2021}. Neither one bootstraps, so neither inherits the error of the last backup, but both fit the value of the behavior policy rather than the optimal one.

\paragraph{Triangle inequality.}
QRL \citep{wang2023qrl} builds the triangle inequality into the network by parameterizing the value as a quasimetric. TDP \citep{kaelbling1993, dhiman2018, jurgenson2020} runs Equation~\ref{eq:transitive} with the hard maximum, which we approximate by drawing subgoal candidates from the dataset. COE \citep{piekos2023} trains a separate generator to propose the subgoal instead of searching for it. TRL \citep{park2026trl} is Equation~\ref{eq:trlloss}, which replaces the maximum with an expectile and takes the subgoal from the state's own trajectory.

\subsection{Datasets}
\label{app:exp-datasets}

Every task and dataset comes from OGBench \citep{park2025ogbench} and we use them as released.

The teleport mazes are the stochastic group. They are the size of the large maze and they hold teleporters. Entering one sends the agent to one of three exits at random, and one exit is a dead end. So the same action has several successors, and the base case of Section~\ref{sec:method-why} fails on them. 

The stitch datasets hold short trajectories, each at most four cell units long, and a task needs up to eight of them joined together. So most state and goal pairs never lie on a common trajectory, which is the coverage problem of Proposition~\ref{prop:reachability}.

The remaining ten tasks are deterministic. Their navigate datasets come from a noisy expert that repeatedly reaches random goals. The play datasets of cube, scene and puzzle come from scripted policies that move objects around without a task in mind.

Every task uses its \textit{-oraclerep-v0} variant \citep{park2026trl}, where the goal is a fixed low-dimensional slice of the state. Table~\ref{tab:datasets} gives the specification of each task.

\begin{table}[h]
\centering
\caption{\textbf{Dataset specifications.} Transitions, trajectories and dimensions of the nineteen tasks. A dataset trajectory is not as long as an evaluation episode, and the last column is the step budget the agent is given at evaluation time.}
\label{tab:datasets}
\footnotesize\ttfamily
\setlength{\tabcolsep}{3.5pt}
\resizebox{\textwidth}{!}{%
\begin{tabular}{llccccccc}
\toprule
Environment & Dataset & Transitions & Trajectories & Traj. length & Obs. dim & Goal dim & Action dim & Episode length \\
\midrule
pointmaze-teleport-navigate    & \multirow{4}{*}{Teleport}  & 1M & 1000 & 1000 & 2 & 2 & 2 & 1000 \\
pointmaze-teleport-stitch      &                            & 1M & 5000 & 200 & 2 & 2 & 2 & 1000 \\
antmaze-teleport-navigate      &                            & 1M & 1000 & 1000 & 29 & 2 & 8 & 1000 \\
antmaze-teleport-stitch        &                            & 1M & 5000 & 200 & 29 & 2 & 8 & 1000 \\
\cmidrule(lr){1-9}
pointmaze-large-stitch         & \multirow{5}{*}{Stitch}    & 1M & 5000 & 200 & 2 & 2 & 2 & 1000 \\
antmaze-large-stitch           &                            & 1M & 5000 & 200 & 29 & 2 & 8 & 1000 \\
humanoidmaze-medium-stitch     &                            & 2M & 5000 & 400 & 69 & 2 & 21 & 2000 \\
humanoidmaze-large-stitch      &                            & 2M & 5000 & 400 & 69 & 2 & 21 & 2000 \\
antsoccer-arena-stitch         &                            & 1M & 5000 & 200 & 42 & 2 & 8 & 1000 \\
\cmidrule(lr){1-9}
pointmaze-large-navigate       & \multirow{10}{*}{Deterministic} & 1M & 1000 & 1000 & 2 & 2 & 2 & 1000 \\
antmaze-large-navigate         &                            & 1M & 1000 & 1000 & 29 & 2 & 8 & 1000 \\
humanoidmaze-medium-navigate   &                            & 2M & 1000 & 2000 & 69 & 2 & 21 & 2000 \\
humanoidmaze-large-navigate    &                            & 2M & 1000 & 2000 & 69 & 2 & 21 & 2000 \\
antsoccer-arena-navigate       &                            & 1M & 1000 & 1000 & 42 & 2 & 8 & 1000 \\
cube-single-play               &                            & 1M & 1000 & 1000 & 28 & 3 & 5 & 200 \\
cube-double-play               &                            & 1M & 1000 & 1000 & 37 & 6 & 5 & 500 \\
scene-play                     &                            & 1M & 1000 & 1000 & 40 & 7 & 5 & 750 \\
puzzle-3x3-play                &                            & 1M & 1000 & 1000 & 55 & 9 & 5 & 500 \\
puzzle-4x4-play                &                            & 1M & 1000 & 1000 & 83 & 16 & 5 & 500 \\
\bottomrule
\end{tabular}}
\end{table}

\subsection{Implementation Details}
\label{app:exp-impl}

GTRL is built on the Transitive RL implementation \citep{park2026trl}. Critic is an ensemble of two networks. Each returns a logit and the value is its sigmoid, so $Q$ stays in $(0,1)$ where Equation~\ref{eq:value} puts it. The loss $L$ inside Equation~\ref{eq:expectile} is the binary cross entropy between the prediction and the target, as in TRL. The target critic $\bar{Q}$ is a Polyak average of the critic at rate $\eta$. The policy is trained against the smaller of the two ensemble values, and Equation~\ref{eq:ddpgbc} divides that value by its own absolute mean, so $\alpha_{\mathrm{BC}}$ does not have to be retuned when the scale of $Q$ moves.

The composition $\bar{Q}(s,a,w)\,\bar{Q}(w,a_w,g)$ needs the subgoal $w$ to lie in the same space as the state. The \textit{oraclerep} tasks break this, since the goal is a slice of a state and not a state. We follow TRL and keep two critics. The first is conditioned on full states and carries the loss of Equation~\ref{eq:gtrlloss}. The second is conditioned on the oracle goal and is fit to the first by regression. Only the second is used for policy extraction.

Equation~\ref{eq:hw} reuses $\bar{Q}(s',a',g)$, which Equation~\ref{eq:td} has already computed, so the weight costs one further forward pass for $\bar{Q}(s,a,g)$ and nothing else. Both terms come from the target critic, so no gradient flows through the weight. We clip it and then normalize it to mean one over the decomposable goals of the batch.

\subsection{Training and Evaluation}
\label{app:exp-training}

We train every agent for $10^{6}$ gradient steps on four seeds. Each OGBench task has five evaluation goals and we run 50 episodes for each. A run is scored by its success rate over all five goals, averaged over the checkpoints at 800K, 900K and 1M, following OGBench \citep{park2025ogbench}. Every table reports the mean and the standard deviation of that score over the four seeds.

\subsection{Hyperparameters}
\label{app:exp-hparams}

Table~\ref{tab:hp-shared} lists the hyperparameters that are the same on every task. Table~\ref{tab:hp-task} lists the three that are tuned per task, which are the distance weight $\lambda$ of Equation~\ref{eq:trlloss}, the behavioral cloning coefficient $\alpha_{\mathrm{BC}}$ of Equation~\ref{eq:ddpgbc}, and the discount factor $\gamma$. Baseline hyperparameters are left as they were in the \citep{park2026trl} default.

\begin{table}[h]
\centering
\caption{\textbf{Shared hyperparameters.} These are the same on all nineteen tasks.}
\label{tab:hp-shared}
\footnotesize\ttfamily
\begin{tabular}{ll}
\toprule
Hyperparameter & Value \\
\midrule
Gradient steps & $10^{6}$ \\
Optimizer & Adam \\
Learning rate & $3 \times 10^{-4}$ \\
Batch size & $1024$ \\
MLP size & $[512, 512, 512]$ \\
Nonlinearity & GELU \\
Layer normalization & True \\
Target network update rate $\eta$ & $0.005$ \\
Subgoal expectile $\kappa$ ($m = 1$) & $0.7$ \\
One-step expectile $\kappa$ ($m = 0$) & $0.5$ \\
Hindsight weight clip $c$ & $1.0$ \\
Value goal ratio $(p_{\mathrm{cur}}, p_{\mathrm{traj}}, p_{\mathrm{rand}})$ & $(0,\, 0.5,\, 0.5)$ \\
Value goal geometric sampling & True \\
Actor goal ratio $(p_{\mathrm{cur}}, p_{\mathrm{traj}}, p_{\mathrm{rand}})$ & $(0,\, 1,\, 0)$, and $(0,\, 0.5,\, 0.5)$ on the five stitch tasks \\
Actor goal geometric sampling & False \\
Policy extraction & Reparameterized gradients \\
Oracle distillation & True \\
Evaluation episodes per goal & $50$ \\
\bottomrule
\end{tabular}
\end{table}

\begin{table}[h]
\centering
\caption{\textbf{Task-specific hyperparameters.} The distance weight $\lambda$, the behavioral cloning coefficient $\alpha_{\mathrm{BC}}$, and the discount factor $\gamma$.}
\label{tab:hp-task}
\footnotesize\ttfamily
\begin{tabular}{llccc}
\toprule
Environment & Dataset & $\lambda$ & $\alpha_{\mathrm{BC}}$ & $\gamma$ \\
\midrule
pointmaze-teleport-navigate & \multirow{4}{*}{Teleport} & 0.1 & 0.1 & 0.99 \\
pointmaze-teleport-stitch & & 0.7 & 10.0 & 0.99 \\
antmaze-teleport-navigate & & 0.0 & 0.25 & 0.99 \\
antmaze-teleport-stitch & & 0.0 & 0.25 & 0.99 \\
\cmidrule(lr){1-5}
pointmaze-large-stitch & \multirow{5}{*}{Stitch} & 0.7 & 0.03 & 0.99 \\
antmaze-large-stitch & & 0.0 & 0.25 & 0.99 \\
humanoidmaze-medium-stitch & & 0.0 & 0.1 & 0.995 \\
humanoidmaze-large-stitch & & 0.0 & 0.1 & 0.995 \\
antsoccer-arena-stitch & & 0.25 & 0.1 & 0.99 \\
\cmidrule(lr){1-5}
pointmaze-large-navigate & \multirow{10}{*}{Deterministic} & 0.1 & 0.1 & 0.99 \\
antmaze-large-navigate & & 0.0 & 0.25 & 0.99 \\
humanoidmaze-medium-navigate & & 0.0 & 0.1 & 0.995 \\
humanoidmaze-large-navigate & & 0.0 & 0.1 & 0.995 \\
antsoccer-arena-navigate & & 0.5 & 0.25 & 0.99 \\
cube-single-play & & 0.4 & 2.5 & 0.99 \\
cube-double-play & & 0.0 & 7.5 & 0.99 \\
scene-play & & 0.7 & 1.5 & 0.99 \\
puzzle-3x3-play & & 0.4 & 1.5 & 0.99 \\
puzzle-4x4-play & & 0.1 & 1.0 & 0.99 \\
\bottomrule
\end{tabular}
\end{table}

\clearpage
\section{Full Results}
\label{app:results}

Table~\ref{tab:full-det} reports the per-goal results behind Table~\ref{tab:standard}. Every task carries five evaluation goals, and the \textit{overall} row of each block is their average.

\begin{table}[h]
\centering
\caption{\textbf{Full results on standard deterministic tasks.} We report success rate (\%) and standard deviation over four seeds, for each of the five evaluation goals of the ten deterministic OGBench \citep{park2025ogbench} tasks. Experiments used \textit{oracle-rep} variants. Best results are marked \textcolor{bestred}{\textbf{RED}}.}
\label{tab:full-det}
\footnotesize\ttfamily
\setlength{\tabcolsep}{2pt}
\resizebox{\textwidth}{!}{%
\begin{tabular}{llcccccccccccc}
\toprule
Environment & Task & BC & FBC & IVL & IQL & TD-n & CRL & MC & QRL & TDP & COE & TRL & \textcolor{bestred}{\textbf{GTRL}} \\
\midrule
\multirow{6}{*}{pointmaze-large-navigate} & task1    & \scell{53}{24} & \scell{98}{3} & \scell{94}{6} & \scell{92}{9} & \scell{74}{13} & \scell{23}{18} & \scell{13}{12} & \scell{31}{28} & \scell{40}{19} & \scell{67}{26} & \scell{52}{19} & \sbest{100}{1} \\
                                   & task2    & \scell{3}{4} & \scell{17}{6} & \scell{0}{0} & \scell{0}{0} & \scell{0}{0} & \scell{42}{28} & \scell{0}{0} & \scell{5}{6} & \scell{0}{0} & \scell{0}{0} & \scell{1}{2} & \sbest{57}{35} \\
                                   & task3    & \scell{17}{9} & \scell{76}{11} & \sbest{100}{0} & \scell{76}{13} & \scell{40}{13} & \scell{67}{14} & \scell{3}{4} & \scell{0}{0} & \scell{4}{4} & \scell{12}{8} & \scell{7}{5} & \sbest{100}{0} \\
                                   & task4    & \scell{23}{10} & \sbest{79}{8} & \scell{3}{5} & \scell{0}{0} & \scell{9}{9} & \scell{13}{18} & \scell{3}{6} & \scell{1}{2} & \scell{56}{12} & \scell{44}{27} & \scell{41}{13} & \scell{42}{15} \\
                                   & task5    & \scell{30}{22} & \scell{87}{5} & \scell{44}{39} & \scell{0}{0} & \scell{31}{8} & \scell{21}{8} & \scell{0}{0} & \scell{0}{0} & \scell{49}{16} & \scell{48}{16} & \scell{64}{14} & \sbest{97}{3} \\
                                   & overall  & \scell{25}{3} & \scell{71}{4} & \scell{48}{10} & \scell{34}{3} & \scell{31}{7} & \scell{33}{7} & \scell{4}{4} & \scell{7}{7} & \scell{30}{5} & \scell{34}{7} & \scell{33}{5} & \sbest{79}{6} \\
\cmidrule(lr){1-14}
\multirow{6}{*}{antmaze-large-navigate} & task1    & \scell{3}{3} & \scell{13}{11} & \scell{11}{13} & \scell{23}{13} & \scell{63}{16} & \scell{87}{9} & \scell{30}{8} & \scell{51}{15} & \scell{6}{5} & \scell{6}{6} & \scell{57}{13} & \sbest{88}{2} \\
                                   & task2    & \scell{22}{12} & \scell{24}{8} & \scell{19}{5} & \scell{44}{20} & \scell{76}{5} & \scell{64}{25} & \scell{47}{8} & \scell{69}{12} & \scell{23}{6} & \scell{26}{11} & \scell{66}{4} & \sbest{79}{5} \\
                                   & task3    & \scell{53}{10} & \scell{48}{6} & \scell{55}{6} & \scell{82}{6} & \scell{87}{6} & \scell{91}{3} & \scell{86}{6} & \sbest{96}{3} & \scell{69}{9} & \scell{67}{4} & \scell{80}{4} & \scell{88}{5} \\
                                   & task4    & \scell{17}{9} & \scell{7}{3} & \scell{4}{3} & \scell{15}{7} & \scell{21}{8} & \sbest{92}{4} & \scell{17}{10} & \scell{57}{16} & \scell{17}{10} & \scell{14}{1} & \scell{8}{8} & \scell{89}{1} \\
                                   & task5    & \scell{17}{5} & \scell{16}{4} & \scell{18}{12} & \scell{20}{7} & \scell{28}{6} & \sbest{90}{6} & \scell{13}{5} & \scell{61}{12} & \scell{17}{4} & \scell{18}{5} & \scell{18}{11} & \scell{88}{3} \\
                                   & overall  & \scell{22}{4} & \scell{22}{4} & \scell{21}{6} & \scell{37}{8} & \scell{55}{6} & \scell{85}{7} & \scell{39}{5} & \scell{67}{8} & \scell{27}{3} & \scell{26}{3} & \scell{46}{5} & \sbest{87}{1} \\
\cmidrule(lr){1-14}
\multirow{6}{*}{humanoidmaze-medium-navigate} & task1    & \scell{10}{7} & \scell{7}{9} & \scell{29}{4} & \scell{31}{9} & \scell{95}{2} & \scell{91}{8} & \scell{79}{13} & \scell{8}{12} & \scell{4}{3} & \scell{6}{4} & \scell{76}{6} & \sbest{97}{0} \\
                                   & task2    & \scell{6}{4} & \scell{8}{5} & \scell{37}{12} & \scell{82}{9} & \scell{94}{1} & \sbest{96}{1} & \scell{88}{5} & \scell{22}{29} & \scell{7}{3} & \scell{9}{2} & \sbest{96}{2} & \scell{93}{2} \\
                                   & task3    & \scell{8}{5} & \scell{15}{5} & \scell{16}{7} & \scell{6}{6} & \scell{8}{7} & \sbest{72}{17} & \scell{1}{1} & \scell{30}{21} & \scell{12}{3} & \scell{13}{3} & \scell{8}{8} & \scell{59}{8} \\
                                   & task4    & \scell{2}{1} & \scell{3}{3} & \scell{0}{0} & \scell{0}{0} & \scell{8}{12} & \scell{8}{7} & \scell{2}{3} & \scell{9}{9} & \scell{0}{0} & \scell{1}{1} & \scell{11}{3} & \sbest{32}{14} \\
                                   & task5    & \scell{11}{5} & \scell{12}{2} & \scell{39}{16} & \scell{63}{12} & \sbest{97}{3} & \scell{93}{4} & \scell{76}{7} & \scell{20}{19} & \scell{14}{4} & \scell{14}{7} & \scell{92}{3} & \sbest{97}{2} \\
                                   & overall  & \scell{7}{2} & \scell{9}{3} & \scell{24}{4} & \scell{36}{3} & \scell{60}{3} & \scell{72}{5} & \scell{49}{3} & \scell{18}{17} & \scell{7}{0} & \scell{9}{2} & \scell{57}{1} & \sbest{75}{4} \\
\cmidrule(lr){1-14}
\multirow{6}{*}{humanoidmaze-large-navigate} & task1    & \scell{1}{1} & \scell{0}{0} & \scell{9}{3} & \scell{10}{1} & \scell{34}{14} & \scell{49}{17} & \scell{6}{5} & \scell{3}{3} & \scell{0}{0} & \scell{0}{0} & \scell{9}{6} & \sbest{53}{12} \\
                                   & task2    & \scell{0}{0} & \scell{0}{0} & \scell{0}{0} & \scell{0}{0} & \scell{0}{0} & \scell{1}{1} & \scell{0}{0} & \scell{0}{0} & \scell{0}{0} & \scell{0}{0} & \scell{0}{0} & \sbest{4}{4} \\
                                   & task3    & \scell{3}{1} & \scell{2}{1} & \scell{3}{2} & \scell{11}{6} & \scell{64}{14} & \scell{63}{25} & \scell{22}{11} & \scell{9}{6} & \scell{3}{4} & \scell{4}{3} & \scell{29}{9} & \sbest{85}{5} \\
                                   & task4    & \scell{3}{3} & \scell{1}{1} & \scell{1}{1} & \scell{2}{1} & \scell{0}{0} & \scell{9}{8} & \scell{1}{1} & \scell{0}{0} & \scell{2}{2} & \scell{4}{2} & \scell{4}{4} & \sbest{42}{14} \\
                                   & task5    & \scell{2}{2} & \scell{2}{1} & \scell{2}{1} & \scell{0}{0} & \scell{0}{0} & \scell{20}{17} & \scell{0}{0} & \scell{1}{1} & \scell{0}{0} & \scell{2}{2} & \scell{0}{0} & \sbest{22}{22} \\
                                   & overall  & \scell{2}{1} & \scell{1}{1} & \scell{3}{1} & \scell{5}{1} & \scell{20}{2} & \scell{28}{5} & \scell{6}{2} & \scell{3}{2} & \scell{1}{1} & \scell{2}{0} & \scell{8}{1} & \sbest{41}{9} \\
\cmidrule(lr){1-14}
\multirow{6}{*}{antsoccer-arena-navigate} & task1    & \scell{7}{4} & \scell{30}{4} & \scell{76}{5} & \scell{86}{6} & \scell{78}{8} & \scell{52}{11} & \scell{71}{4} & \scell{20}{5} & \scell{12}{1} & \scell{6}{5} & \sbest{89}{3} & \scell{88}{3} \\
                                   & task2    & \scell{6}{3} & \scell{24}{7} & \scell{62}{5} & \sbest{92}{4} & \scell{69}{9} & \scell{36}{4} & \scell{49}{11} & \scell{21}{3} & \scell{7}{6} & \scell{11}{2} & \scell{85}{5} & \scell{80}{5} \\
                                   & task3    & \scell{2}{3} & \scell{14}{6} & \scell{82}{8} & \scell{87}{5} & \scell{87}{5} & \scell{49}{10} & \scell{55}{6} & \scell{13}{6} & \scell{1}{2} & \scell{4}{3} & \scell{89}{5} & \sbest{90}{4} \\
                                   & task4    & \scell{2}{1} & \scell{11}{7} & \scell{34}{6} & \sbest{59}{2} & \scell{35}{11} & \scell{13}{7} & \scell{22}{7} & \scell{3}{3} & \scell{3}{1} & \scell{2}{1} & \scell{48}{12} & \scell{49}{6} \\
                                   & task5    & \scell{1}{1} & \scell{6}{3} & \scell{54}{4} & \sbest{63}{7} & \scell{52}{12} & \scell{22}{8} & \scell{12}{5} & \scell{8}{3} & \scell{3}{1} & \scell{3}{1} & \scell{53}{3} & \sbest{63}{7} \\
                                   & overall  & \scell{3}{1} & \scell{17}{2} & \scell{62}{2} & \sbest{77}{3} & \scell{64}{4} & \scell{35}{5} & \scell{42}{3} & \scell{13}{1} & \scell{5}{1} & \scell{5}{2} & \scell{73}{4} & \scell{74}{1} \\
\cmidrule(lr){1-14}
\multirow{6}{*}{cube-single-play}  & task1    & \scell{7}{9} & \scell{17}{5} & \scell{87}{3} & \scell{97}{3} & \scell{91}{5} & \scell{64}{6} & \sbest{98}{2} & \scell{6}{7} & \scell{4}{3} & \scell{7}{7} & \sbest{98}{2} & \scell{93}{5} \\
                                   & task2    & \scell{8}{6} & \scell{18}{12} & \scell{93}{4} & \scell{96}{3} & \scell{96}{4} & \scell{61}{8} & \sbest{99}{1} & \scell{8}{4} & \scell{5}{4} & \scell{6}{12} & \scell{97}{3} & \scell{96}{3} \\
                                   & task3    & \scell{9}{6} & \scell{22}{5} & \scell{93}{2} & \sbest{99}{1} & \scell{98}{1} & \scell{69}{9} & \scell{98}{2} & \scell{9}{3} & \scell{1}{2} & \scell{2}{3} & \sbest{99}{1} & \scell{94}{7} \\
                                   & task4    & \scell{6}{2} & \scell{20}{6} & \scell{85}{7} & \scell{92}{4} & \scell{93}{2} & \scell{56}{10} & \sbest{97}{1} & \scell{2}{2} & \scell{3}{2} & \scell{27}{17} & \scell{93}{6} & \scell{93}{2} \\
                                   & task5    & \scell{5}{5} & \scell{14}{7} & \scell{82}{4} & \scell{89}{5} & \scell{94}{3} & \scell{66}{18} & \sbest{98}{3} & \scell{3}{2} & \scell{0}{0} & \scell{24}{21} & \scell{87}{7} & \scell{88}{6} \\
                                   & overall  & \scell{7}{2} & \scell{18}{5} & \scell{88}{2} & \scell{95}{1} & \scell{95}{2} & \scell{63}{5} & \sbest{98}{1} & \scell{6}{3} & \scell{3}{1} & \scell{13}{8} & \scell{95}{2} & \scell{93}{4} \\
\cmidrule(lr){1-14}
\multirow{6}{*}{cube-double-play}  & task1    & \scell{6}{5} & \scell{17}{8} & \scell{88}{4} & \sbest{92}{6} & \scell{48}{10} & \scell{77}{8} & \scell{40}{8} & \scell{7}{4} & \scell{6}{2} & \scell{1}{2} & \scell{73}{5} & \scell{80}{4} \\
                                   & task2    & \scell{0}{0} & \scell{1}{1} & \scell{78}{5} & \sbest{84}{7} & \scell{0}{0} & \scell{42}{9} & \scell{0}{0} & \scell{0}{0} & \scell{0}{0} & \scell{0}{0} & \scell{23}{7} & \scell{47}{4} \\
                                   & task3    & \scell{0}{0} & \scell{0}{0} & \scell{75}{6} & \sbest{85}{2} & \scell{0}{0} & \scell{39}{10} & \scell{0}{0} & \scell{0}{0} & \scell{0}{0} & \scell{0}{0} & \scell{30}{11} & \scell{38}{4} \\
                                   & task4    & \scell{0}{0} & \scell{1}{2} & \scell{8}{5} & \sbest{12}{9} & \scell{0}{0} & \scell{1}{1} & \scell{0}{0} & \scell{0}{0} & \scell{0}{0} & \scell{0}{0} & \scell{3}{3} & \scell{2}{1} \\
                                   & task5    & \scell{0}{0} & \scell{2}{2} & \sbest{47}{14} & \scell{45}{8} & \scell{1}{1} & \scell{17}{4} & \scell{0}{0} & \scell{0}{0} & \scell{0}{0} & \scell{0}{0} & \scell{18}{7} & \scell{19}{2} \\
                                   & overall  & \scell{1}{1} & \scell{4}{2} & \scell{59}{2} & \sbest{64}{4} & \scell{10}{2} & \scell{35}{3} & \scell{8}{2} & \scell{1}{1} & \scell{1}{0} & \scell{0}{0} & \scell{30}{5} & \scell{37}{1} \\
\cmidrule(lr){1-14}
\multirow{6}{*}{scene-play}        & task1    & \scell{14}{8} & \scell{60}{6} & \scell{97}{3} & \sbest{99}{1} & \scell{96}{4} & \scell{71}{13} & \scell{72}{14} & \scell{19}{8} & \scell{43}{4} & \scell{32}{8} & \scell{97}{2} & \scell{96}{5} \\
                                   & task2    & \scell{2}{1} & \scell{13}{3} & \scell{92}{4} & \sbest{96}{2} & \scell{94}{2} & \scell{14}{4} & \scell{26}{8} & \scell{2}{2} & \scell{9}{2} & \scell{2}{2} & \scell{95}{3} & \sbest{96}{2} \\
                                   & task3    & \scell{2}{3} & \scell{21}{9} & \scell{83}{8} & \scell{89}{8} & \scell{78}{6} & \scell{33}{9} & \scell{25}{15} & \scell{1}{1} & \scell{3}{4} & \scell{2}{3} & \sbest{97}{3} & \scell{92}{2} \\
                                   & task4    & \scell{3}{1} & \scell{19}{8} & \scell{43}{28} & \scell{13}{11} & \scell{79}{7} & \scell{12}{3} & \scell{4}{3} & \scell{6}{3} & \scell{2}{2} & \scell{3}{1} & \scell{76}{17} & \sbest{81}{12} \\
                                   & task5    & \scell{0}{0} & \scell{3}{4} & \scell{23}{16} & \scell{10}{9} & \scell{31}{13} & \scell{2}{2} & \scell{1}{1} & \scell{1}{1} & \scell{0}{0} & \scell{0}{0} & \scell{18}{9} & \sbest{57}{5} \\
                                   & overall  & \scell{4}{2} & \scell{23}{2} & \scell{68}{10} & \scell{61}{3} & \scell{75}{3} & \scell{26}{5} & \scell{25}{5} & \scell{6}{2} & \scell{12}{1} & \scell{8}{2} & \scell{77}{2} & \sbest{84}{4} \\
\cmidrule(lr){1-14}
\multirow{6}{*}{puzzle-3x3-play}   & task1    & \scell{4}{4} & \scell{10}{1} & \scell{4}{2} & \scell{99}{1} & \scell{99}{1} & \scell{15}{8} & \scell{93}{12} & \scell{3}{3} & \scell{5}{1} & \scell{7}{3} & \scell{99}{1} & \sbest{100}{0} \\
                                   & task2    & \scell{1}{2} & \scell{1}{1} & \scell{2}{2} & \scell{99}{1} & \scell{99}{1} & \scell{6}{3} & \scell{90}{10} & \scell{0}{0} & \scell{1}{1} & \scell{2}{2} & \scell{99}{1} & \sbest{100}{0} \\
                                   & task3    & \scell{1}{1} & \scell{1}{1} & \scell{2}{1} & \scell{99}{1} & \scell{99}{1} & \scell{1}{1} & \scell{69}{15} & \scell{0}{0} & \scell{0}{0} & \scell{1}{2} & \sbest{100}{0} & \scell{99}{1} \\
                                   & task4    & \scell{0}{0} & \scell{1}{1} & \scell{1}{1} & \scell{98}{2} & \scell{99}{1} & \scell{1}{1} & \scell{67}{15} & \scell{0}{0} & \scell{2}{3} & \scell{1}{1} & \scell{98}{1} & \sbest{100}{1} \\
                                   & task5    & \scell{1}{1} & \scell{2}{2} & \scell{1}{1} & \scell{95}{2} & \scell{99}{1} & \scell{2}{2} & \scell{83}{14} & \scell{0}{0} & \scell{1}{1} & \scell{0}{0} & \scell{99}{1} & \sbest{100}{0} \\
                                   & overall  & \scell{1}{1} & \scell{3}{1} & \scell{2}{1} & \scell{98}{0} & \scell{99}{0} & \scell{5}{1} & \scell{81}{12} & \scell{1}{1} & \scell{2}{0} & \scell{2}{0} & \scell{99}{0} & \sbest{100}{0} \\
\cmidrule(lr){1-14}
\multirow{6}{*}{puzzle-4x4-play}   & task1    & \scell{1}{1} & \scell{1}{1} & \scell{6}{4} & \scell{33}{14} & \scell{9}{7} & \scell{1}{1} & \scell{7}{2} & \scell{0}{0} & \scell{1}{1} & \scell{0}{0} & \scell{47}{5} & \sbest{76}{8} \\
                                   & task2    & \scell{0}{0} & \scell{1}{1} & \scell{4}{4} & \scell{0}{0} & \scell{6}{3} & \scell{0}{0} & \scell{3}{4} & \scell{0}{0} & \scell{0}{0} & \scell{1}{1} & \sbest{17}{5} & \scell{13}{6} \\
                                   & task3    & \scell{0}{0} & \scell{1}{1} & \scell{6}{1} & \scell{58}{10} & \scell{4}{4} & \scell{1}{1} & \scell{4}{3} & \scell{0}{0} & \scell{1}{1} & \scell{0}{0} & \scell{38}{13} & \sbest{66}{9} \\
                                   & task4    & \scell{1}{1} & \scell{1}{1} & \scell{6}{6} & \scell{22}{5} & \scell{8}{4} & \scell{1}{1} & \scell{4}{3} & \scell{0}{0} & \scell{0}{0} & \scell{0}{0} & \scell{34}{2} & \sbest{65}{9} \\
                                   & task5    & \scell{0}{0} & \scell{0}{0} & \scell{6}{1} & \scell{29}{9} & \scell{3}{3} & \scell{0}{0} & \scell{7}{4} & \scell{0}{0} & \scell{1}{1} & \scell{0}{0} & \scell{32}{6} & \sbest{65}{9} \\
                                   & overall  & \scell{0}{0} & \scell{1}{0} & \scell{5}{2} & \scell{28}{4} & \scell{6}{3} & \scell{0}{0} & \scell{5}{2} & \scell{0}{0} & \scell{0}{0} & \scell{0}{0} & \scell{34}{4} & \sbest{57}{8} \\
\bottomrule
\end{tabular}}
\end{table}

\section{Additional Ablations: Split Expectile}
\label{app:ablations}

\begin{figure}[h]
\centering
\includegraphics[width=\textwidth]{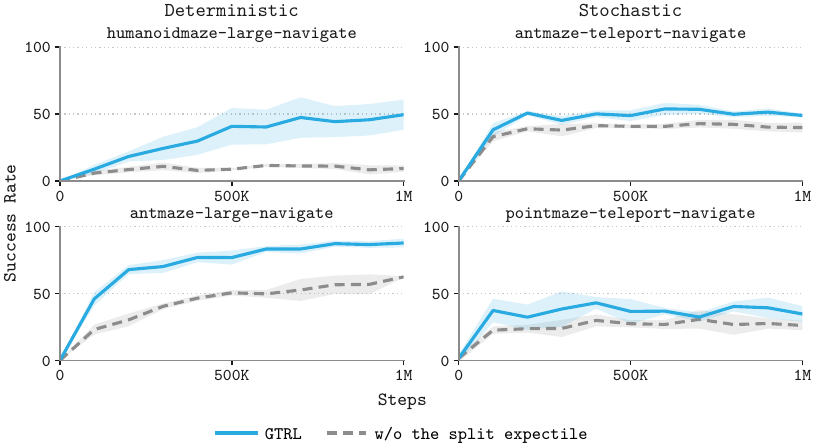}
\caption{Ablation study on \textbf{the split expectile.}}
\label{fig:abl-expectile}
\end{figure}

Equation~\ref{eq:kappa} fits a decomposable goal at $\kappa = 0.7$, which is optimistic over subgoals, and the rest at $\kappa = 0.5$, which is the mean over outcomes. We ablate the split by setting $\kappa = 0.7$ on both. Figure~\ref{fig:abl-expectile} shows a loss of 40.2 points on \textit{humanoidmaze-large-navigate} and 25.3 points on \textit{antmaze-large-navigate}, with smaller drops of 9.0 and 8.6 points on the stochastic tasks. Half of the batch carries the one-step target, so fitting it optimistically inflates values everywhere.

\end{document}